\documentclass{article} 
\usepackage{iclr2025_conference,times}

\usepackage{hyperref}

\usepackage{url}
\usepackage{lineno}  
\usepackage{algorithm}
\usepackage{algorithmic}
\usepackage{graphicx}
\usepackage{amsmath}
\usepackage{cleveref}
\usepackage{amssymb}
\usepackage{booktabs}
\usepackage{csquotes}
\usepackage{graphicx}  
\usepackage[numbers]{natbib}
\usepackage{subcaption}  

\definecolor{mydarkblue}{RGB}{0, 0, 139}

\hypersetup{colorlinks=true, citecolor=mydarkblue,linkcolor=mydarkblue}
\usepackage{longtable}

\title{Curvature Informed Furthest Point Sampling}

\author{Shubham Bhardwaj\textsuperscript{$^{\dagger}$$^{\star}$},~
    Ashwin Vinod\textsuperscript{$^{\dagger}$$^{\diamond}$},~
    Soumojit Bhattacharya\textsuperscript{$^{\ddagger}$$^{\diamond}$},~
    Aryan Koganti\textsuperscript{$^{\ddagger}$},\\[0.2em]
    \textbf{Aditya Sai Ellendula\textsuperscript{$^{\dagger}$}},~
    \textbf{Balakrishna Reddy\textsuperscript{$^{\S}$}}\\[0.5em]
    \textsuperscript{$^{\dagger}$}University of Texas at Austin,~
    \textsuperscript{$^{\ddagger}$}Indian Institute of Technology, Kharagpur,~
    \textsuperscript{$^{\S}$}Jio Platforms
}

\footnotetext[1]{\textsuperscript{$^{\star}$}Corresponding author: shubham.bhardwaj@utexas.edu}
\footnotetext[2]{\textsuperscript{$^{\diamond}$}Equal contribution}

\iclrfinalcopy 

\begin{document}

\vspace{-.2cm}
\maketitle

\vspace{-.2cm}
\begin{abstract}
  Point cloud representation has gained traction due to its efficient memory usage and simplicity in acquisition, manipulation, and storage. However, as  point cloud sizes increase, effective down-sampling becomes essential to address the computational requirements of downstream tasks. Classical approaches, such as furthest point sampling (FPS), perform well on benchmarks but rely on heuristics and overlook geometric features, like curvature, during down-sampling.

In this paper, We introduce a reinforcement learning-based sampling algorithm that enhances FPS by integrating curvature information. Our approach ranks points by combining FPS-derived soft ranks with curvature scores computed by a deep neural network, allowing us to replace a proportion of low-curvature points in the FPS set with high-curvature points from the unselected set. Existing differentiable sampling techniques often suffer from training instability, hindering their integration into end-to-end learning frameworks. By contrast, our method achieves stable end-to-end learning, consistently outperforming baseline models across multiple downstream geometry processing tasks. We provide comprehensive ablation studies, with both qualitative and quantitative insights into the effect of each feature on performance. Our algorithm establishes state-of-the-art results for classification, segmentation and shape completion, showcasing its robustness and adaptability.
\end{abstract}

\begin{figure}[h]
    \centering
    \begin{subfigure}[b]{0.3\textwidth}
        \centering
        \includegraphics[width=\textwidth, angle=0]{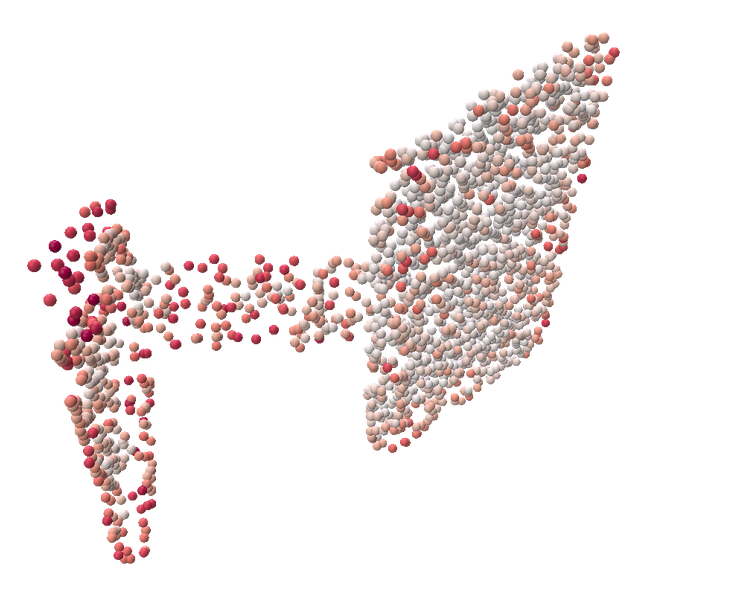}
        \caption{Ground Truth}
        \label{fig:image1}
    \end{subfigure}
    \hfill
    \begin{subfigure}[b]{0.3\textwidth}
        \centering
        \includegraphics[width=\textwidth, angle=0]{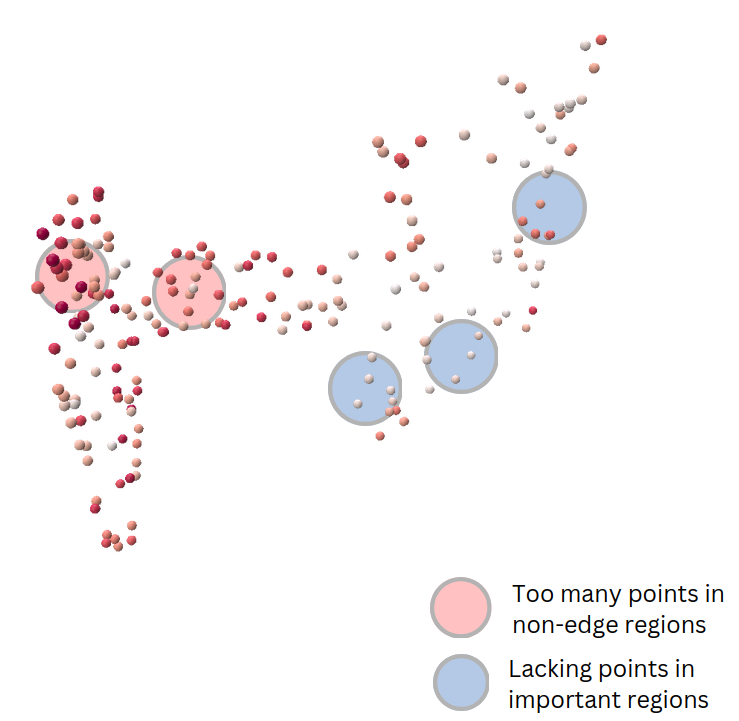}
        \caption{Downsampled using FPS}
        \label{fig:image2}
    \end{subfigure}
    \hfill
    \begin{subfigure}[b]{0.3\textwidth}
        \centering
        \includegraphics[width=\textwidth, angle=0]{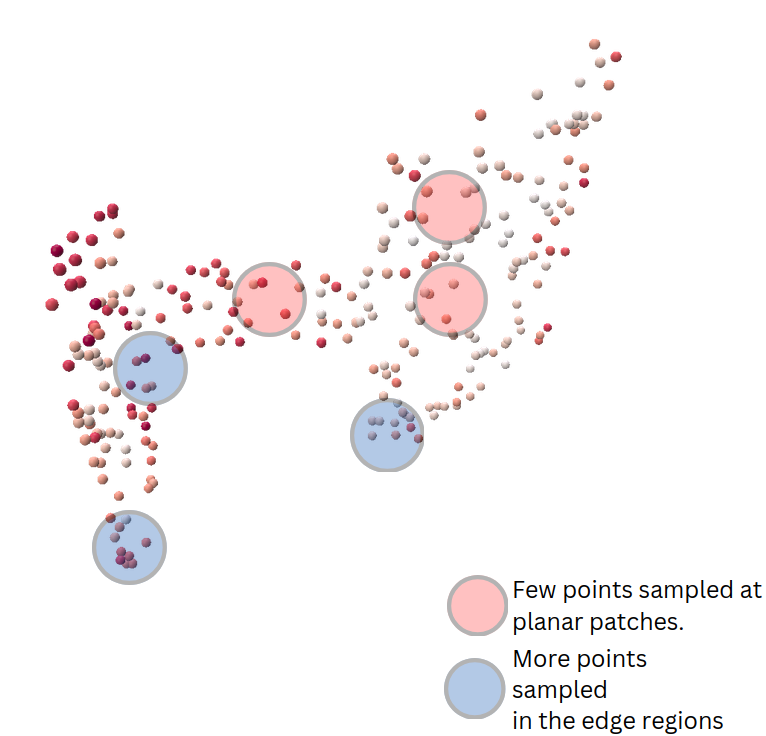}
        \caption{Downsampled using CFPS}
        \label{fig:image3}
    \end{subfigure}

    \caption{2048 points of a monitor downsampled to 256 points showing CFPS retains more structural information than
traditional FPS. The scalar map shows curvature values increasing from white to red. Notice more higher curvature points with CFPS.}
    \label{fig:three_images}
\end{figure}

\section{Introduction}

Applications in robotics, autonomous driving, and augmented reality have seen increased reliance on 3D data, with point clouds being a common representation. A point cloud represents a 3D scene as a collection of colored points in space, capturing both the structure and visual details of the environment. However, working with large-scale point clouds can be computationally expensive, necessitating techniques to reduce their size.
The reduction in the size is generally done via down-sampling. An effective sampling operation is expected to preserve local shape details and the overall shape structure of the point cloud.
			
The main challenge in down-sampling is selecting which data points to retain in a way that maintains the original geometry of the point cloud. One straightforward and efficient approach is to sample the $k$ farthest points from the existing points in the cloud, where $k$ is the desired number of points for the down-sampled representation. This method is known as Furthest Point Sampling (FPS). \cite{pan2021variational}, \cite{xue2023ulip}.

A major limitation of the Furthest Point Sampling (FPS) algorithm is that it is task-agnostic. As a result, it cannot prioritize points that are highly relevant to specific downstream tasks. For instance, in tasks such as 3D reconstruction, points with high curvature but low density are particularly important. To draw an analogy, in 2D image processing, max pooling selects the most activated feature from a local spatial grid. Similarly, in 3D point clouds, we need an algorithm that can perform a similar function. However, the local spatial grid in 3D is defined by the K-nearest neighbor relationships between point features, and the farthest point cannot be selected directly because the point features exist in a higher-dimensional space. Therefore, FPS is not directly applicable in feature spaces.

A simple formulation would model the subset creation problem of point clouds via MLP layers. The main issue with this simple formulation is that end-to-end learning becomes highly unstable for downstream tasks, since we may lose the entire structure of the object while creating a random subset of points. This approach works well for discriminative tasks \cite{wu_2023_attention}, \cite{nezhadarya2020adaptive}.
Therefore, recent approaches have followed a two-step approach. The first step involves learning a model with the furthest point sampling as an intermediate point cloud subset creation layer. The second step involves freezing all the layers of the trained model and adding new MLP layers to replace the farthest point sampling layer. In this step, the aim is to train the sampling layers only. The two-step procedure is useful in practice and outperforms the traditional furthest-point sampling approach.  The drawback of this approach is the overhead that comes along with re-training the model. This drawback hinders the mass adoption of such 2-step approaches.

We introduce curvature-informed furthest point sampling (CFPS) that is robust enough to allow end-to-end learning over downstream tasks while outperforming both traditional and differentiable formulations of the point cloud sub-sampling. The first step in CFPS involves generating a soft rank for each point based on when it enters the furthest point set. This soft rank is then combined with curvature scores obtained from PCPNet \cite{guerrero2018pcpnet} to create a joint rank. Finally, points in the furthest point set with the lowest joint rank are replaced by points that were initially excluded but have a high joint rank. By integrating curvature with the FPS feature, CFPS ensures that the selected point subsets effectively capture both local and global geometric characteristics of the shape during the early stages of training. The key advantage of this approach is its ability to outperform traditional FPS by using a policy gradient-based estimator to dynamically adjust the exchange ratio, selecting the optimal number of points to swap.
We summarize our contributions as follows : 
\begin{itemize}

\item We propose a sub-sampling algorithm that is compatible with the end-to-end learning paradigm.
\item We extend the furthest point sampling algorithm with traditional and layer-specific features for stable task-specific subset selection.
\item We propose the first downsampling algorithm that outperforms previous end-to-end learning based downsampling methods on both generative and discriminative tasks.

\end{itemize}

Please Note that works such as SampleNet \cite{lang2020samplenet}, and Learning to Sample \cite{dovrat2019learning} do not conform to our paradigm for comparison. Our method and the baselines proposed are for an end-to-end learning model wherein down-sampling occurs multiple times throughout the network. The approaches \cite{lang2020samplenet}\cite{dovrat2019learning} proposed are for a 2-step training approach where sampling only happens once. i.e. before sending the input point cloud to the network. Our method achieves state-of-the-art performance on shape reconstruction and classification with pre-trained models.
\section{Related work}

\subsection{Deep Learning with Point Clouds} Deep learning has transformed 3D data analysis, driving advancements in object detection \cite{lang2019pointpillars}, segmentation \cite{qi2017pointnet}, and registration \cite{yang2020teaser}. PointNet and PointNet++ laid the groundwork for point cloud processing, inspiring architectures like DGCNN \cite{wang2019dynamicgraphcnnlearning}, PointCNN \cite{li2018pointcnn}, and PointConv \cite{wu2020pointconvdeepconvolutionalnetworks}, which improve feature extraction by leveraging local geometric structures and spatial properties.

Point clouds offer key advantages: \textbf{ease of capture} via 3D scanners or photogrammetry \cite{liu2019deep}, \textbf{memory efficiency} compared to voxel grids, \textbf{high fidelity} shape representation critical for reconstruction \cite{pan2021variational}, and \textbf{broad applicability} across navigation, perception, and depth estimation tasks \cite{wang2021attention}.

Recent methods, such as Atrous Convolution \cite{pan2020pointatrousgraph}, KPConv \cite{thomas2019kpconv}, and Graph Attention Networks (GAT) \cite{velickovic2017graph}, further refine point cloud processing. To handle large datasets efficiently, down-sampling techniques like \textbf{Furthest Point Sampling (FPS)} reduce computational costs while preserving diverse geometric features essential for segmentation and detection.

\subsection{Furthest Point Sampling (FPS)}


The \textbf{Furthest Point Sampling (FPS)} algorithm \cite{eldar1997farthest} starts with an initial point selected arbitrarily and iteratively adds the point that is furthest from all previously selected points. This process continues until the desired number of points are selected. The primary goal of FPS is to achieve maximum coverage of the shape by selecting points that are well spread out across the point cloud.

\subsubsection{Formulation of FPS}

Given a set of points \( P = \{p_1, p_2, \dots, p_n\} \) in a Euclidean space, FPS selects a subset \( S \subset P \) of size \( k \), where each point in \( S \) is iteratively chosen to be the farthest point from the set of already selected points. The selection process is defined as:

\begin{enumerate}
    \item Initialize \( S = \{p_1\} \), where \( p_1 \) is arbitrarily chosen.
    \item For each subsequent point \( p_i \), compute the distance from all points in \( S \), and select \( p_i \) as the point with the maximum minimum distance to any point in \( S \):
    \[
    p_i = \arg \max_{p_j \in P \setminus S} \min_{p_s \in S} \| p_j - p_s \|
    \]
    \item Repeat until the desired number of points \( k \) are selected.
\end{enumerate}
\subsubsection{Limitations of FPS}

The algorithm aims to achieve maximum coverage of the shape, but FPS is task-agnostic \cite{qi2017pointnet++} \cite{yu2018pu} \cite{li2018pointcnn}. Despite its practical success, FPS has several other limitations:

\begin{itemize}
    \item \textbf{Non-differentiability}: FPS is not differentiable, which makes it difficult to incorporate into end-to-end, task-specific learning frameworks.
    \item \textbf{Feature Space Incompatibility}: FPS is designed for Euclidean spaces, making it less suitable for high-dimensional feature spaces commonly used in deep learning models for point cloud processing.
    \item \textbf{Sensitivity to Outliers}: FPS can be sensitive to outliers, as they can significantly affect the selection of farthest points, leading to suboptimal point subsets.
\end{itemize}

\subsection{Critical Points Layer}
The Critical Points Layer (CPL) \cite{nezhadarya2020adaptive} is a differentiable down-sampling module that works directly with the high-dimensional point-level features. The critical points layer passes the most active features to the next layer. 

The downside of directly working in the feature space is that during the initial few epochs of the training stage, the down-sampling layer would make decisions based solely on the features that have not yet been generalized to the downstream task. This observation suggests that the network would take many epochs to converge, as highlighted in the paper.

\subsection{Attention-based point cloud edge sampling (APES)}
The APES algorithm \cite{wu_2023_attention} is a down-sampling method that combines neural network-based learning and mathematical statistics-based direct point selecting. APES focuses on outline points (edge points) for sampling point cloud subsets for downstream tasks. It uses the attention mechanism to compute correlation maps and sample edge points whose properties are reflected in these correlation maps.

\section{Proposed Solution}
We propose a novel down-sampling algorithm that builds upon Furthest Point Sampling (FPS). In our approach, we use FPS to create a soft feature that ranks each point based on its membership score to the furthest point set. This ranking serves as a measure of a point's relevance in the point cloud.

Our method draws inspiration from point cloud sampling techniques such as \cite{metzer2021self}, which rely on proxies for the sparseness and sharpness of local regions to determine whether a point is worthy of inclusion in the sample set. Intuitively, points that are sparsely distributed in planar regions are sufficient to capture the overall shape, whereas regions with more intricate details or higher curvature—such as the armrest of a chair as visualized—require a denser sampling to accurately preserve the shape's local geometry. FPS based features target preserving the global structure while curvature based features target preserving the local structure.

\subsection{Curvature}
Curvature is a fundamental geometric property that characterizes how a surface bends or deviates from being flat at a given point. In the context of point clouds, curvature provides crucial information about the local shape and features of the underlying surface.


Prior methods have incorporated curvature to identify distinct features for classification and segmentation purposes, to preserve important geometric features, for adaptive sampling, and applications like edge detection. By calculating local curvature, they can differentiate between flat, smoothly curved, or sharp regions, which is crucial for effective point cloud analysis.

In this paper, we opt for mean curvature as our curvature measure due to its practical significance in capturing the local surface variations that are highly relevant for shape preservation in down-sampling. 

\textbf{Mean Curvature $(H)$}: The average of the two principal curvatures \cite{rusinkiewicz2004estimating}, $H = \frac{k_1 + k_2}{2}$

Specifically, we estimate the normals at each point using MSECNet \cite{xiu2023msecnet}, which is a fast and robust normal estimator that utilizes multi-scale patch based features to predict the normal value for each point in the point cloud.

\subsection{Algorithm}
Consider a input point cloud $X \in \mathbb{R}^{B \times N \times 3}$, where $B$ is the batch size $N$ is number of points in the point cloud. Let $X = X_{core} \cup \bar {X}_{core}$, where $X_{core}$ is the furthest point set of points and $\bar X_{core}$ is the non furthest point set. 
We obtain $X_{core}$ by running the furthest point sampling algorithm.
The modification we introduce to the furthest point sampling algorithm is to remove low curvature points from the furthest point set $X_{core} \in \mathbb{R}^{B \times K \times 3}$ and replace it with high curvature points from the non-furthest point set $\bar X_{core}\in \mathbb{R}^{B \times (N-K) \times 3}$. Intuitively, sampling additional points from a planar region is redundant and does not add value. We replace points with low curvature in \( X_{\text{core}} \) with high curvature points from \( \bar{X}_{\text{core}} \), since points with low curvature (typically on flat regions) are redundant and provide limited value in reconstructing sharp features. Let curvature values $C \in \mathbb{R}^{B \times N}$. We then compute the ranking order of points via the furthest point sampling algorithm $\mathbf{F} \in \{0, \ldots, N-1\}^{B \times N}$. One of the changes that we have added to the furthest point sampling algorithm is the conversion of ranking order to a normalized ranking order or soft rank $\mathbf{S}$. $\mathbf{S} = \frac{\mathbf{F}}{N-1}$

The soft rank $\mathbf{S}$ converts furthest point ranks into normalized features that can be used in conjunction with other features such as curvature values to create a joint rank that honors the furthest point heuristic and curvature. The joint rank for each point is calculated as $\mathbf{J} = \mathbf{C}*\mathbf{S}$

Finally, the points are organized into the furthest point set called $X_{core}$ and the points discarded by furthest point sampling called $\bar X_{core}$. Given the joint rank \( \mathbf{J} \), we determine the points to exchange between \( X_{\text{core}} \) and \( \bar{X}_{\text{core}} \). The number of points to be exchanged is controlled by a learned \textbf{exchange ratio} \( G \in [0, 1]^{B \times 1} \), which indicates the percentage of total points \( N \) to swap between the two sets. The exchange ratio is learned via a stochastic policy network / ratio estimator $\pi_\phi: C \rightarrow G$. where $G$ is the percentage over total input points $N$ to be exchanged between $X_{core}$ and $\bar {X}_{core}$. Our policy network $\pi_\phi$ is a temporal convolutional network that can accept arbitrary size input point cloud curvature values $C$.

The actions of our policy network are modeled using a Beta distribution, which is ideal for continuous outputs constrained to the interval \([0, 1]\). 

There are three key strategies for computing gradients through stochastic nodes: Control variates, Score functions  (REINFORCE rule) and the reparameterization trick \cite{mohamed2020monte}. In our approach, we incorporate all three techniques to reduce the variance of the gradient estimates.

For control variates, we use a dynamic baseline $b$ to reduce the variance in the gradient estimates, as outlined in  \cref{alg:reinforce}. Specifically, we use an exponential moving average (EMA) of the past rewards to update the baseline.



The action $a_t \in [0, 1]$ at each time step $t$ is sampled from this Beta distribution: $a_t \sim Beta(\alpha_t, \beta_t)$. The log probability of the sampled action is then used during training to update the policy network based on the reward received. This allows the network to optimize the exchange ratio and improve the down-sampling strategy over time. The log-probabilities are used in the loss function during the policy gradient update: $$\mathcal{L}_{policy} = -\mathbb{E}_{\pi_\phi}[(R_t - b_t)\ \log(\pi_\phi(a_t|s_t))]$$ where $R_t$ is the reward at time $t$, $b_t$ is the baseline (choice of which is explained later) and $\pi_{\phi}(a_t|s_t)$ is the policy's action probability.

We describe the key components of our Markov Decision process (MDP), denoted as $(\mathcal{S}, \mathcal{A}, \mathcal{P}, \mathcal{R})$ where: 

\begin{itemize}
   \item \textbf{State Space:} \( \mathcal{S} \) comprises curvature values \( C_t \in \mathbb{R}^{B \times N} \).
    \item \textbf{Action Space:} \( \mathcal{A} \) is the exchange ratio \( G_t \in [0, 1]^B \), where each \( G_t^i \) determines the fraction of points swapped for the \( i \)-th batch.
    \item \textbf{Transition Function:} \( \mathcal{P}(s_{t+1} | s_t, a_t) \) assumes deterministic transitions, \( s_{t+1} = C_{t+1} \).
    
    \item \textbf{Reward:} The reward \(\mathcal{R}(s_t, a_t)\) at time step \(t\) is a scalar value that indicates how well the action $a_t$ (exchange ratio $G_t)$ performs in terms of the down-sampling objective. The reward is derived from the loss function \(\mathcal{L}_\theta(X, G_t)\), where \(X\) is the input point cloud, and \(G_t\) is the selected exchange ratio. The reward is noisy and depends on the current network’s parameters \(\theta\).

\end{itemize}

\begin{figure*}
    \includegraphics[width=\textwidth]{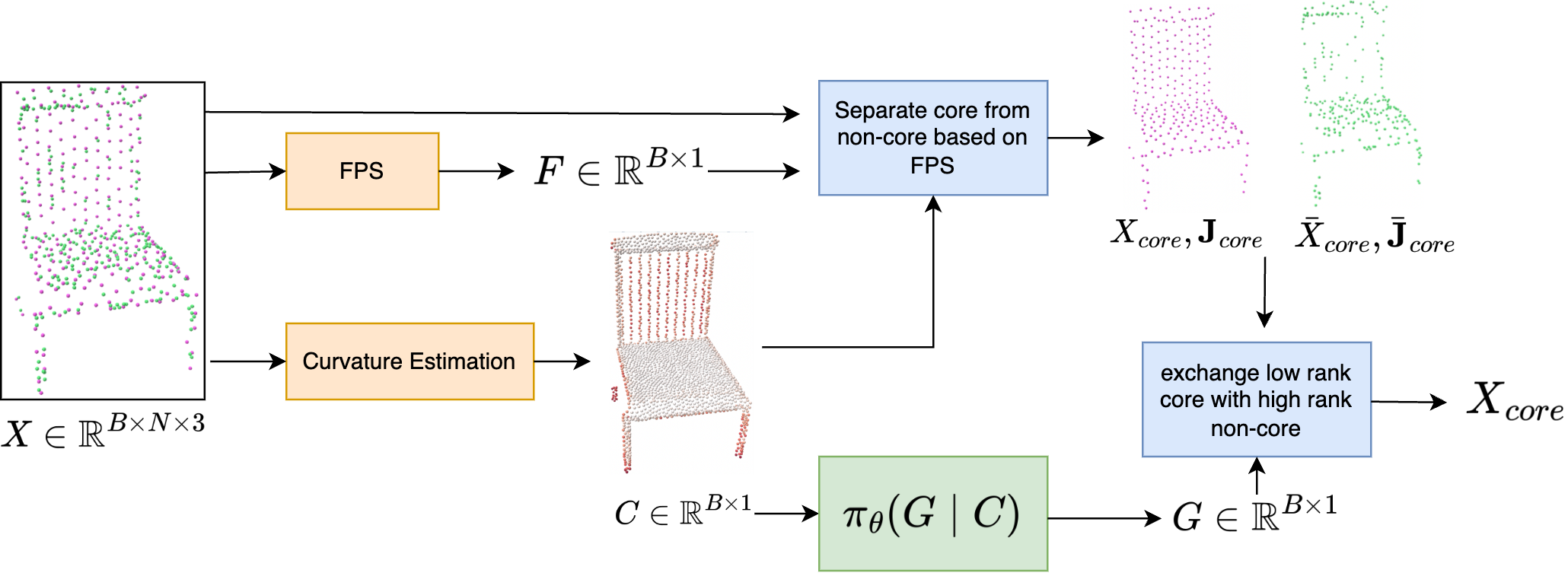}
    \caption{Architecture diagram of the Curvature-aware Furthest Point Sampling (CFPS) pipeline. The pipeline involves swapping the least important points from the core set ($X_{core}$), selected by the Furthest Point Sampling (FPS) algorithm, with the most important points from the unselected points ($\bar{X}_{core}$). The importance of each point is determined by a joint ranking based on both curvature and FPS rank. The number of points to be exchanged between the core and non-core sets is determined by an exchange ratio predicted by a policy network.}

    \label{fig:arch_diag}
\end{figure*}

 The pipeline for the algorithm is given in Figure \ref{fig:arch_diag} and the complete algorithm is presented in Algorithm \ref{alg:cfps}.
 For theoretical convergence bounds on our REINFORCE algorithm refer Supplimentary.


\begin{algorithm}[htp]
\caption{REINFORCE rule for learning the optimal exchange ratio (called after every min-batch training step of the neural network $\mathcal{M}_\theta$) using the downsampling module}
\label{alg:reinforce}
\begin{algorithmic}[1]
\STATE \textbf{Returns:} $\pi_\phi$ updated policy network\;
\STATE \textbf{Input:} $C \in \mathbb{R}^{T \times B \times N}$ The input features (e.g., curvature) where $T$ is the total number of downsampling blocks, $B$ is the batch size, and $N$ is the number of input features, $\lambda = 0.99$;
\STATE Compute the policy parameters $\alpha, \beta$ using the policy network $\pi_\phi(C)$;
\STATE Sample the reparameterized action $G \sim \text{Beta}(\alpha, \beta)$ using the reparameterization trick;
\STATE Observe reward $R$ that is sampled from the noisy reward function: $R = -L_{\theta}(X, G)$;
\STATE Update the baseline $b$ using an exponential moving average:
$b = \lambda * b + ( 1 - \lambda ) * R$;
\STATE Update the parameters of the ratio prediction module (policy $\pi_\phi$) using the REINFORCE rule:
$\nabla_{\phi} F(\phi) = \left( R - b \right) \nabla_{\phi} \log \pi_{\phi}(G \mid C)$;
\end{algorithmic}
\end{algorithm}



\begin{algorithm}[htp]
\caption{Curvature-Aware Furthest Point Sampling (CFPS)}
\label{alg:cfps}
\begin{algorithmic}[1]
\STATE \textbf{Returns:} $X_{core}$ downsampled point cloud\;
\STATE \textbf{Input:} $X \in \mathbb{R}^{B \times N \times 3}$ Input point cloud, $K$ target size, $\pi_\phi$ policy network\;
\STATE Compute point-wise mean curvature $C \in \mathbb{R}^{B \times N}$ using normals from \cite{xiu2023msecnet} for all points;
\STATE Sample exchange ratio $G \sim \pi_\phi(G \mid C)$ where $G \in [0, 1]^B$;
\STATE Initialize FPS ranking order $\mathbf{F} \in \{0, \ldots, N-1\}^{B \times N}$ using the FPS algorithm;
\STATE Compute normalized rank $\mathbf{S} = \frac{\mathbf{F}}{N-1}$;
\STATE Compute a combined ranking $\mathbf{J} = \mathbf{C} + \mathbf{S}$ (combining curvature and FPS ranks);
    \STATE Obtain core and non-core point sets:
    \STATE \hspace{1em} $X_{core} \gets$ FPS$(X, K)$;
    \STATE \hspace{1em} $\bar{X}_{core} \gets X \setminus X_{core}$;
    \STATE Predict exchange ratio using policy network:
    \STATE \hspace{1em} $G \gets \pi_\phi(C)$;
    \STATE \hspace{1em} $n_{exchange} \gets \lfloor G \cdot N \rfloor$;
    \STATE Select lowest-ranked points from core set:
    \STATE \hspace{1em} $P_{core}^{low} \gets$ select the $n_{exchange}$ points with the lowest rank $\mathbf{J}$ in $X_{core}$;
    \STATE Select highest-ranked points from non-core set:
    \STATE \hspace{1em} $P_{noncore}^{high} \gets$ select the $n_{exchange}$ points with the highest rank $\mathbf{J}$  in $\bar{X}_{core}$;
    \STATE Swap points between core and non-core sets:
    \STATE \hspace{1em} $X_{core} \gets (X_{core} \setminus P_{core}^{low}) \cup P_{noncore}^{high}$;
\RETURN $X_{core}$
\end{algorithmic}
\end{algorithm}

\section{Experiment Results}
We explored two discriminative tasks (Classification and Segmentation) and one generative task (Single-View Partial Point Completion). Details of the segmentation task are provided in the supplementary material.

\subsection{Single-View Partial Point Cloud Completion}
Scanned 3D point clouds are often incomplete due to occlusions, noise, and limited viewpoint of the 3D scanner. This incomplete data hampers the usability of the acquired 3D point cloud data for various tasks, especially for in generative tasks such as shape completion. One challenge in point-cloud based shape completion is the difficulty in preserving detailed topological features of the generated object, which are critical for high-fidelity completed shapes \cite{wang2021voxel}. In our experiments, we demonstrate that the \textbf{sampling procedure} plays a fundamental role in improving the quality and fidelity of the completed shape, particularly when leveraging our CFPS-based sampling strategy.

\subsubsection{Dataset}

We evaluate our method on the \textbf{MVP dataset}\cite{pan2021multi}, large-scale multi-view partial point cloud dataset consisting of over 100,000 high-quality 3D scans. For each 3D CAD model, the dataset contains rendered partial 3D shapes from 26 uniformly distributed camera poses. The training set contains 62,400 partial-complete point cloud pairs and the Test set has 41,800 pairs. Each point cloud contains 2,048 points, making this dataset highly suitable for evaluating point cloud completion at scale

\begin{figure*}[htp]
    
    \begin{tabular}{lllll}
        & Desk & Chair & Monitor & Dressing Table \\ \hline
        Original Point Cloud & 
        \includegraphics[width=0.13\linewidth]{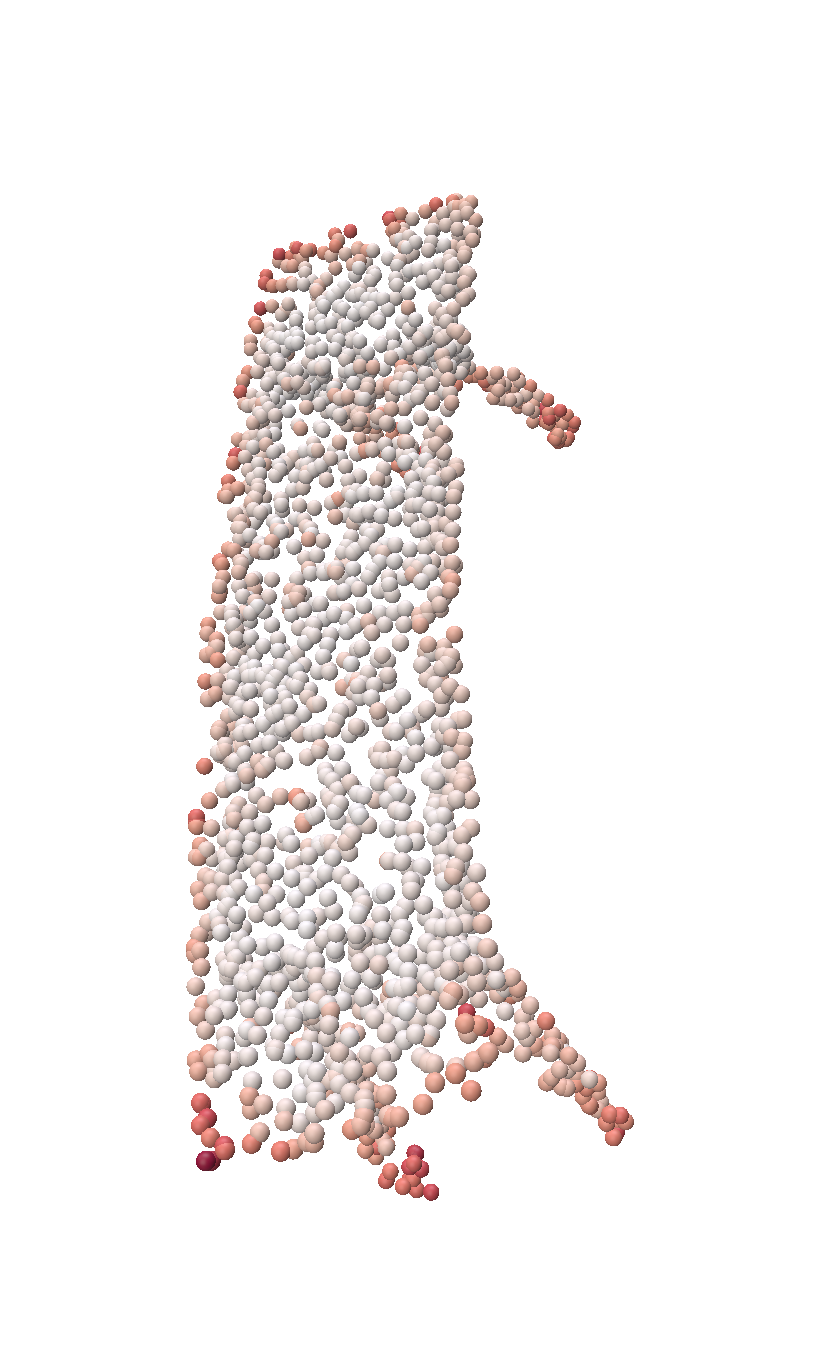} & 
        \includegraphics[width=0.22\linewidth]{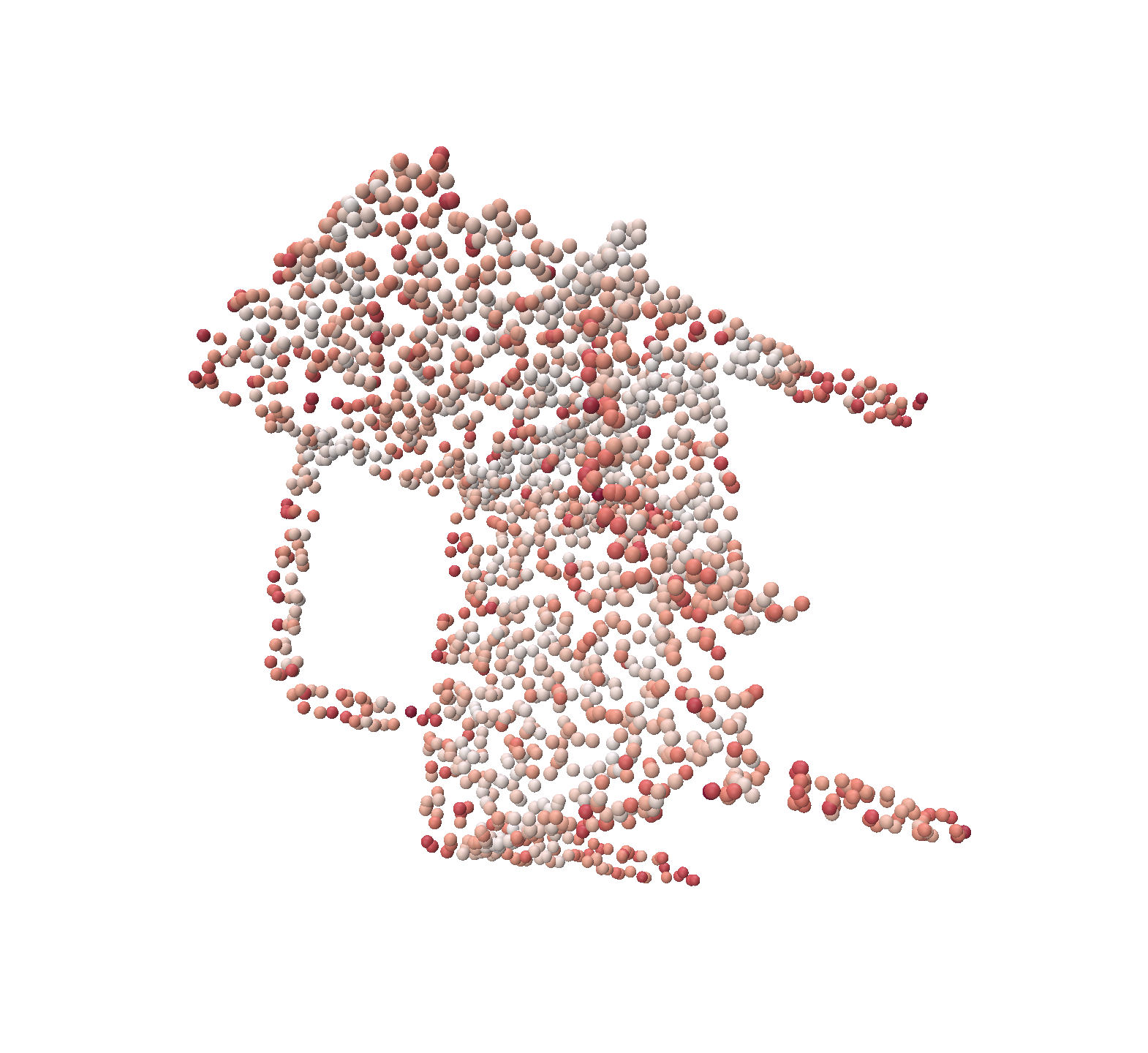} &
        \includegraphics[width=0.22\linewidth]{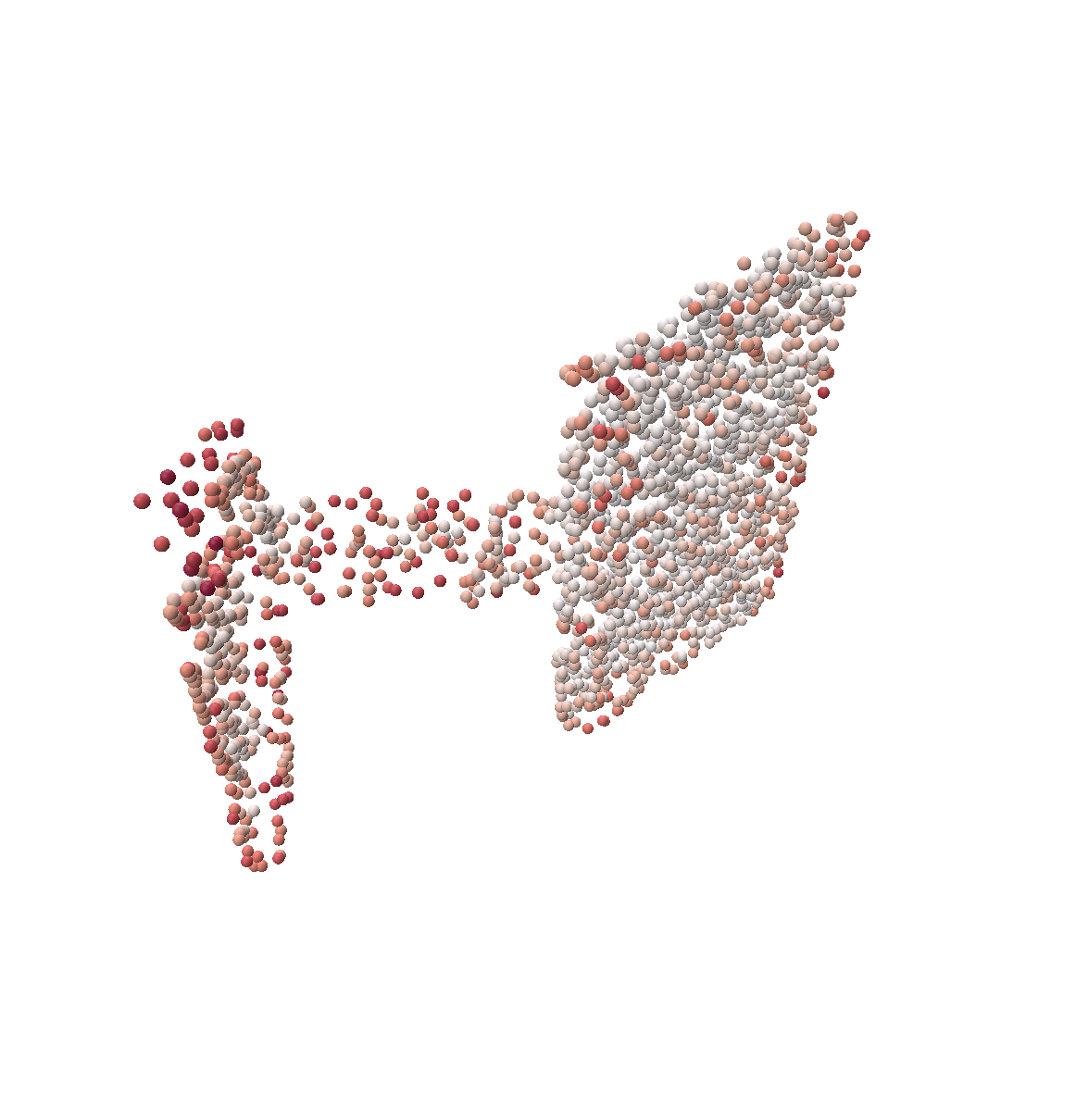} &
        \includegraphics[width=0.18\linewidth]{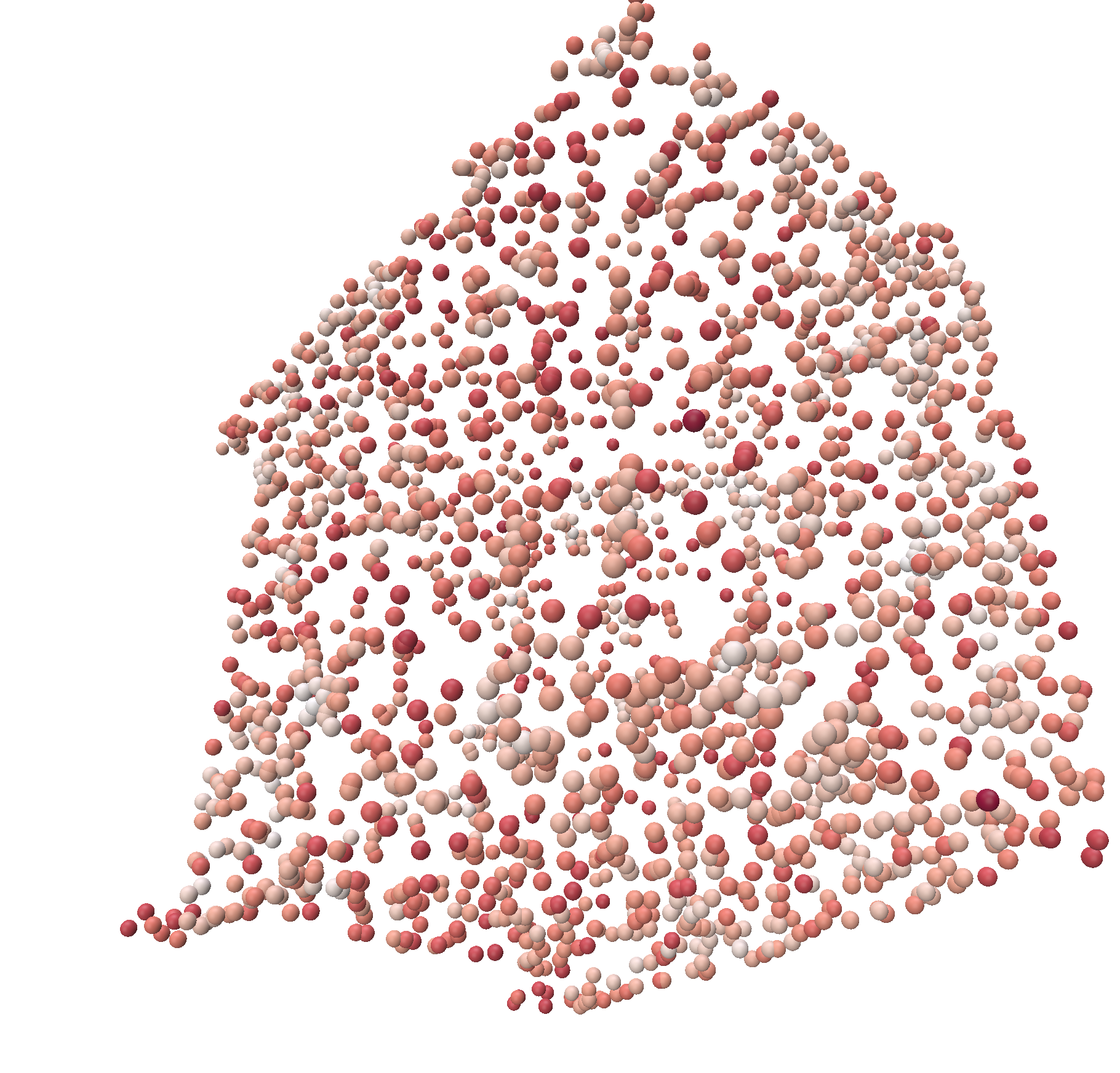} \\
        
        FPS & 
        \includegraphics[width=0.13\linewidth]{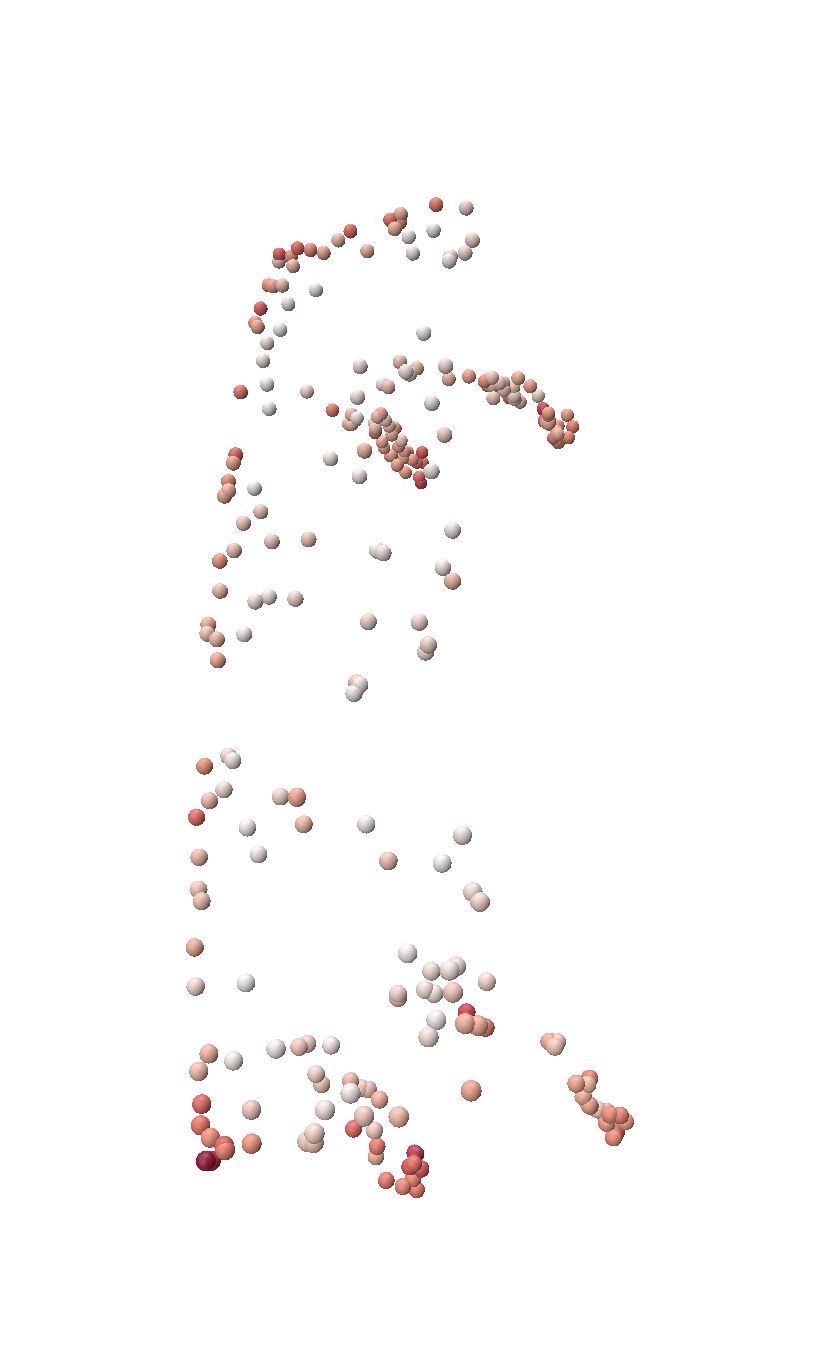} & 
        \includegraphics[width=0.22\linewidth]{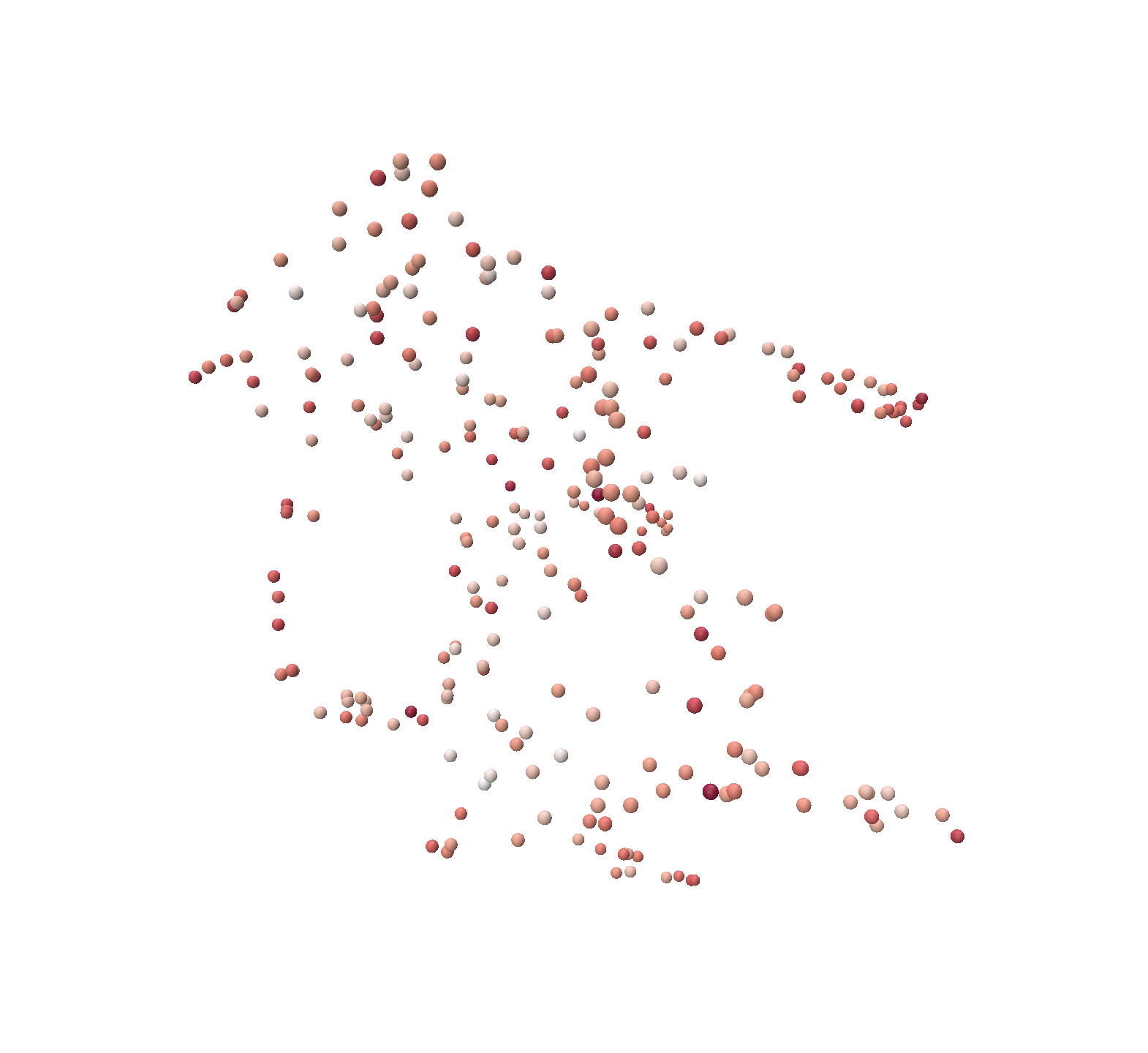} &
        \includegraphics[width=0.22\linewidth]{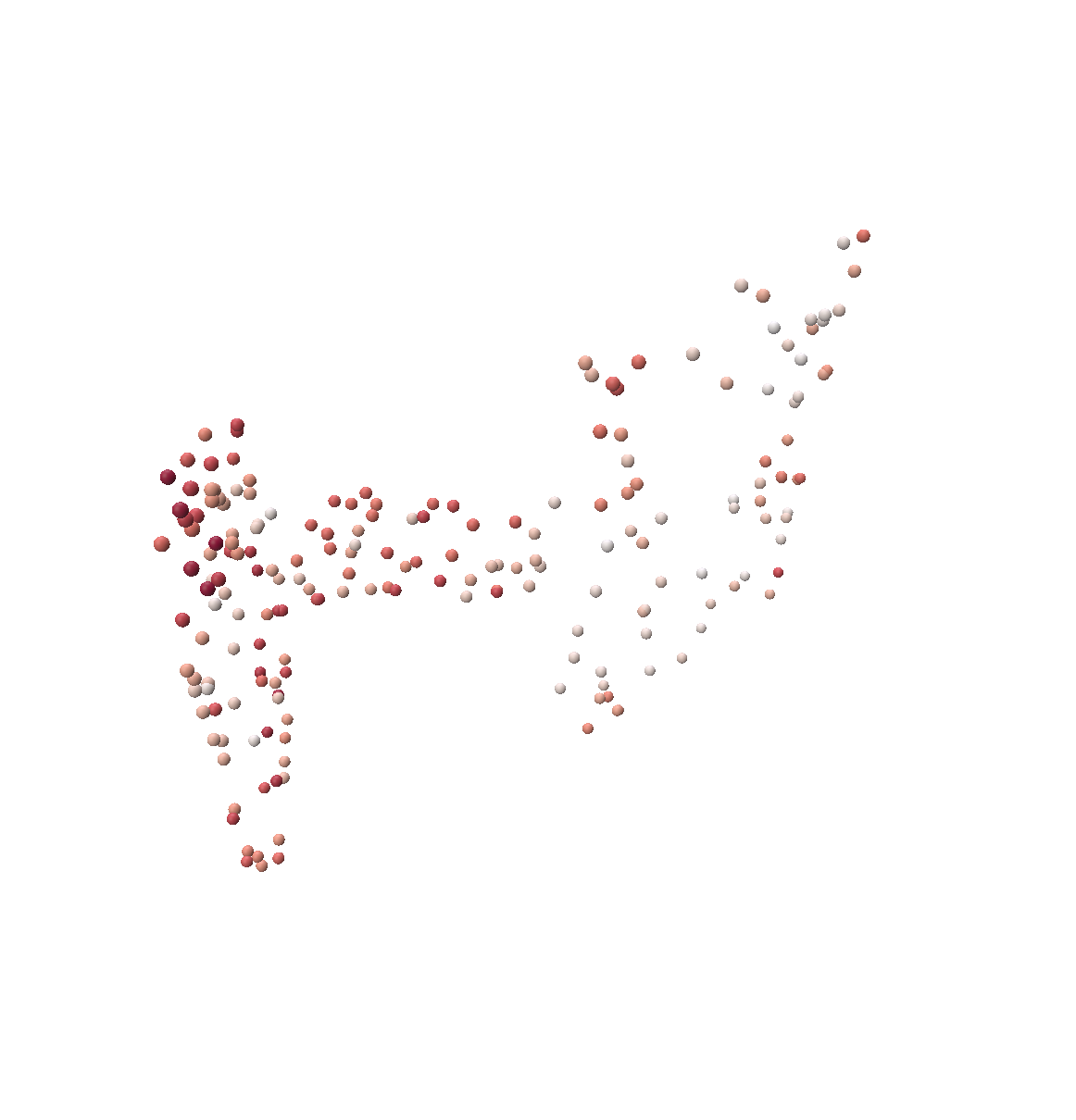} &
        \includegraphics[width=0.18\linewidth]{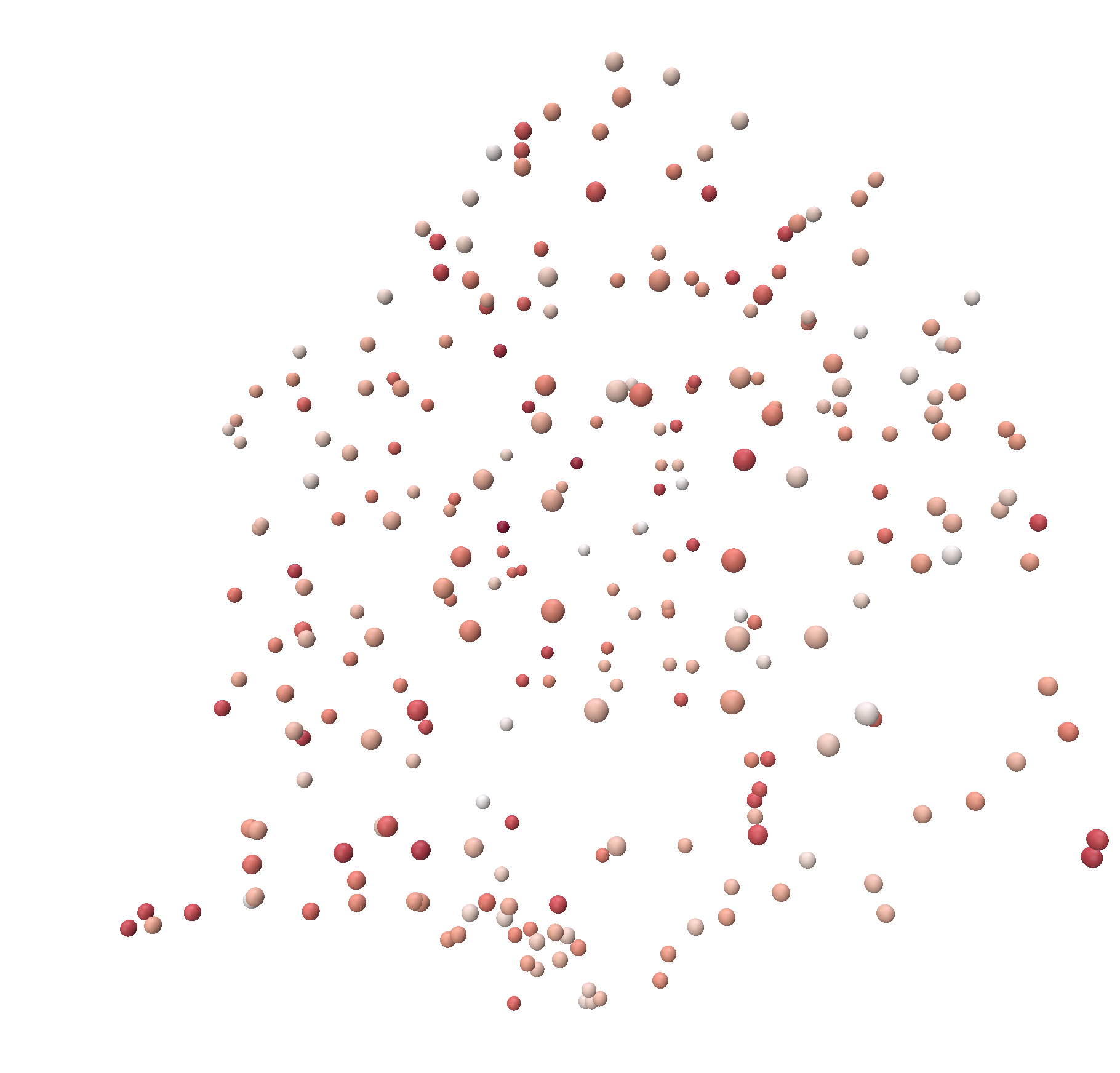} \\
        
        APES Local & 
        \includegraphics[width=0.13\linewidth]{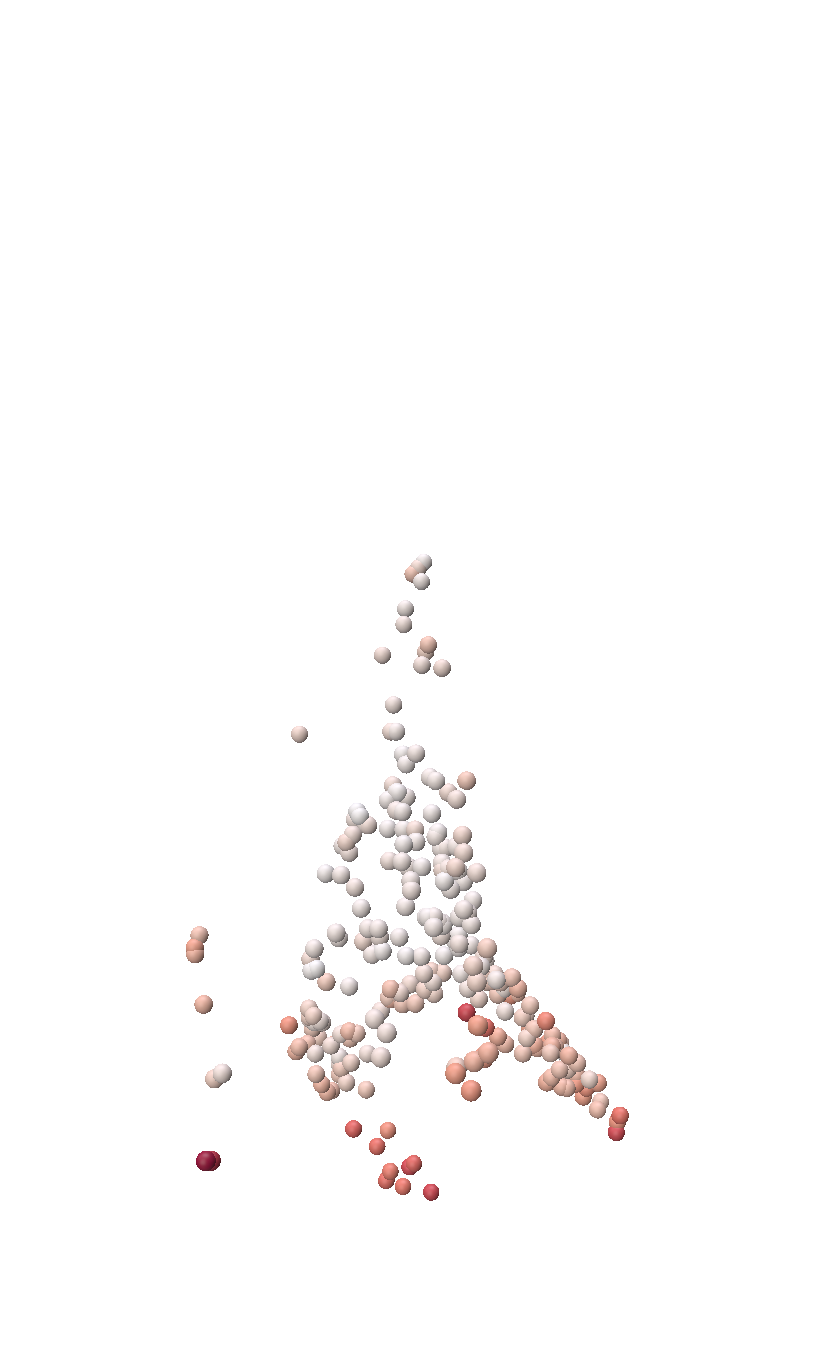} & 
        \includegraphics[width=0.22\linewidth]{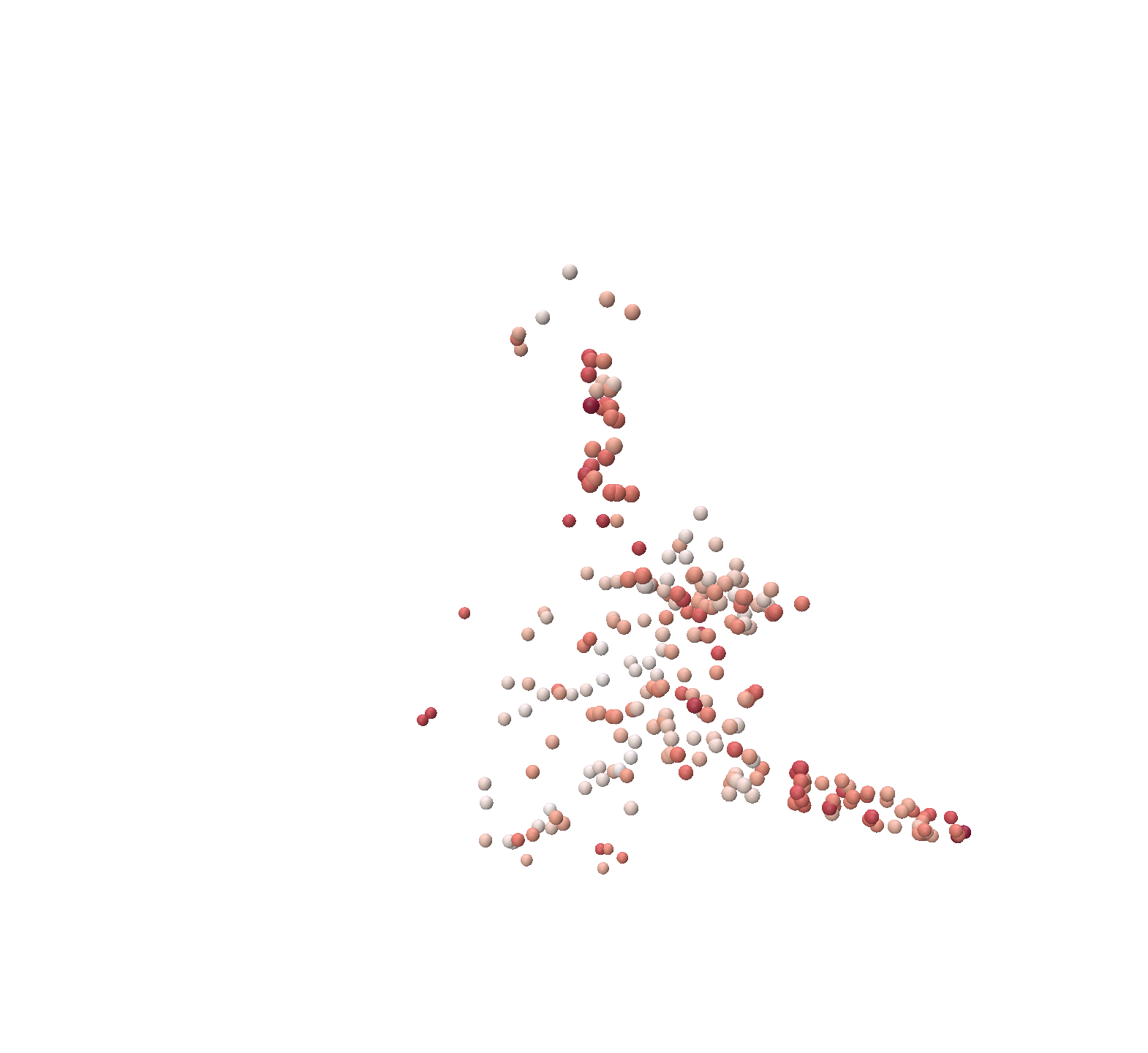} &
        \includegraphics[width=0.22\linewidth]{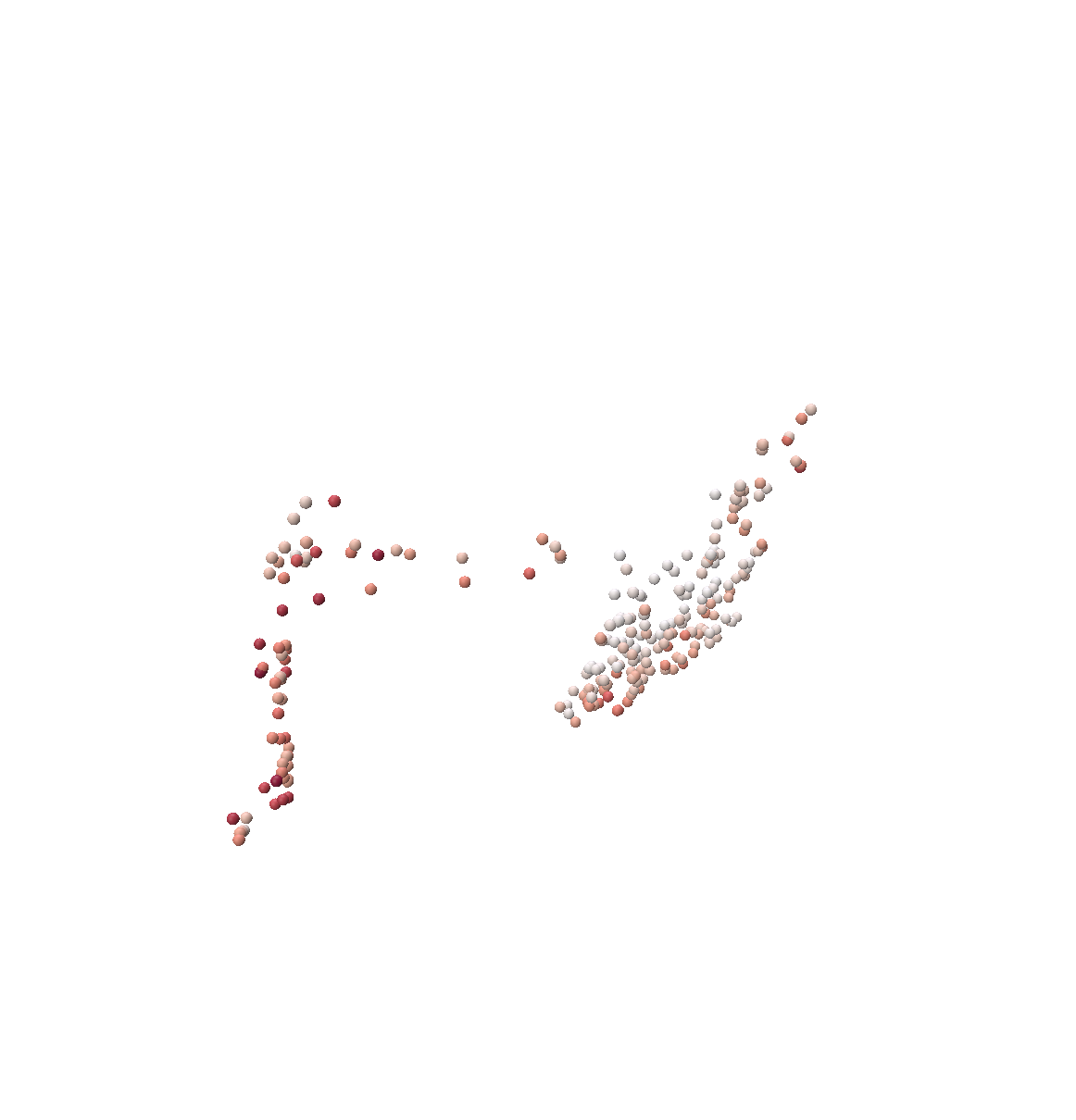} &
        \includegraphics[width=0.18\linewidth]{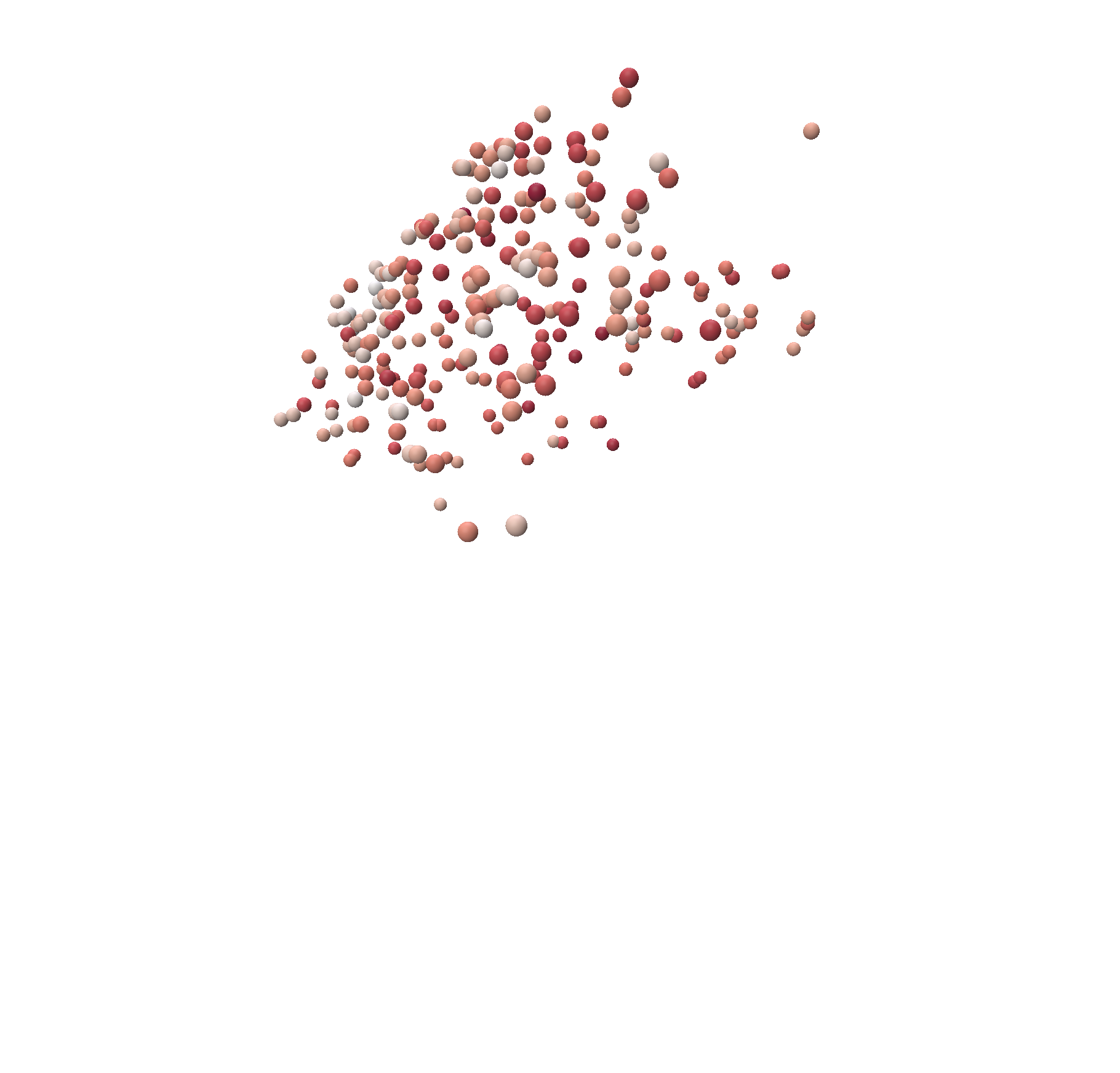} \\
        
        APES Global & 
        \includegraphics[width=0.13\linewidth]{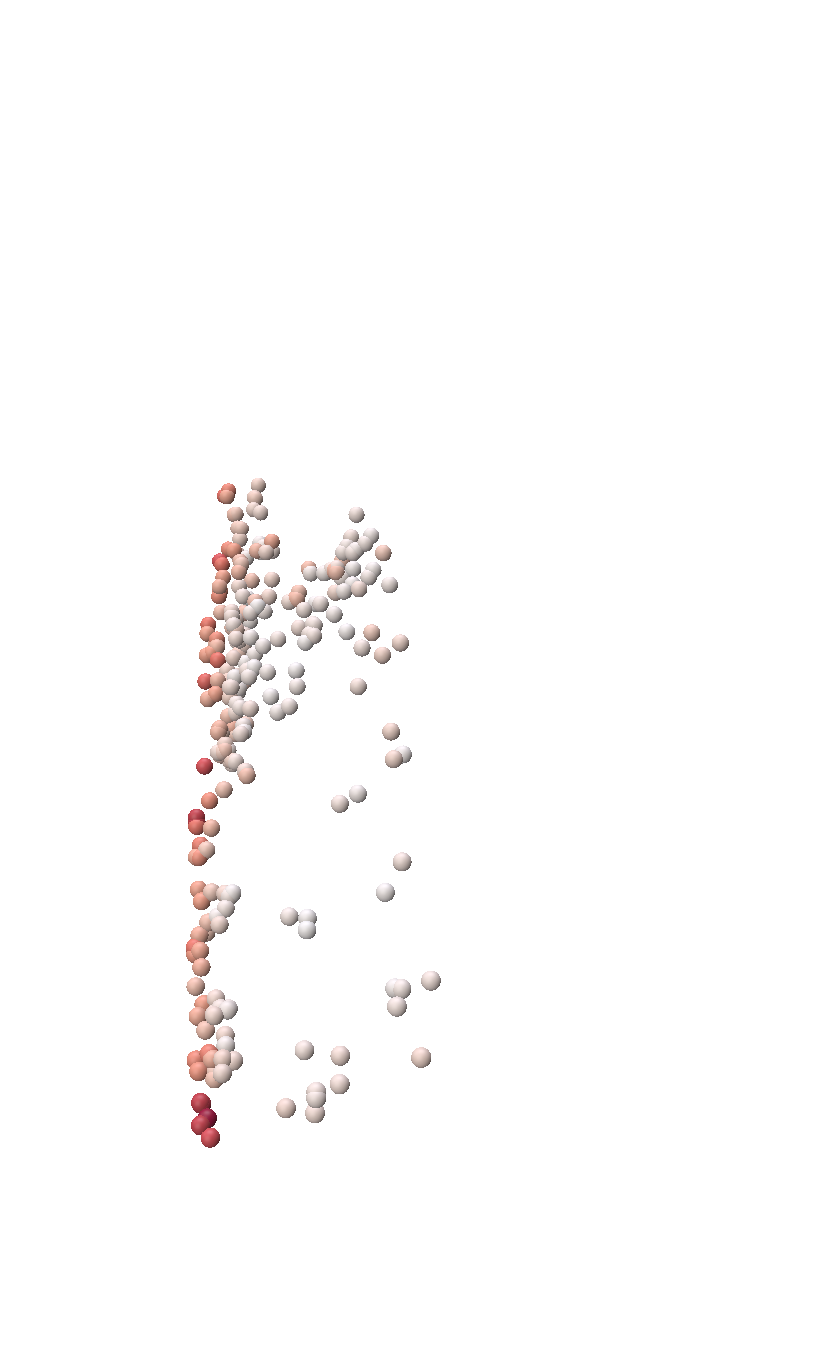} & 
        \includegraphics[width=0.22\linewidth]{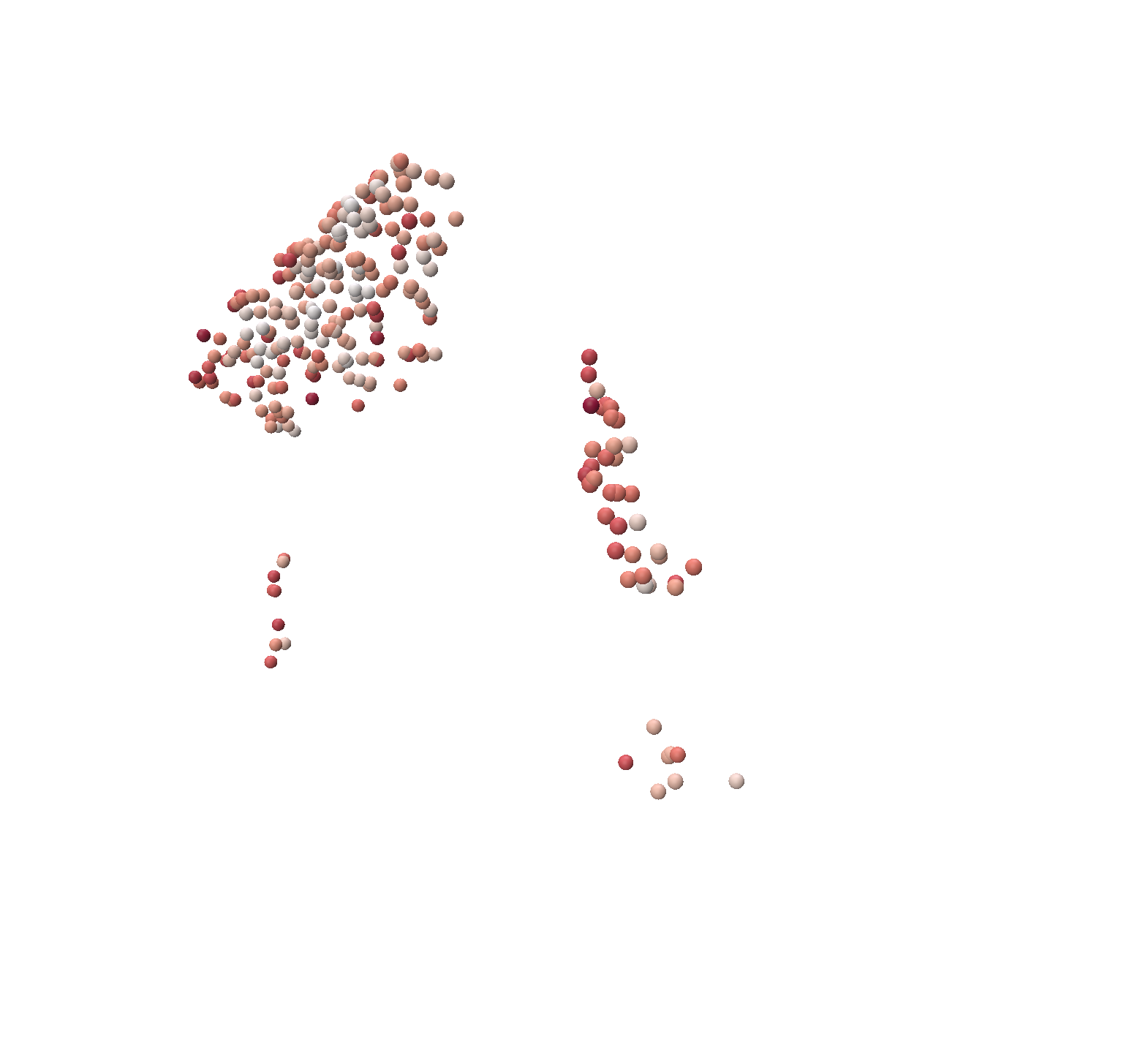} &
        \includegraphics[width=0.22\linewidth]{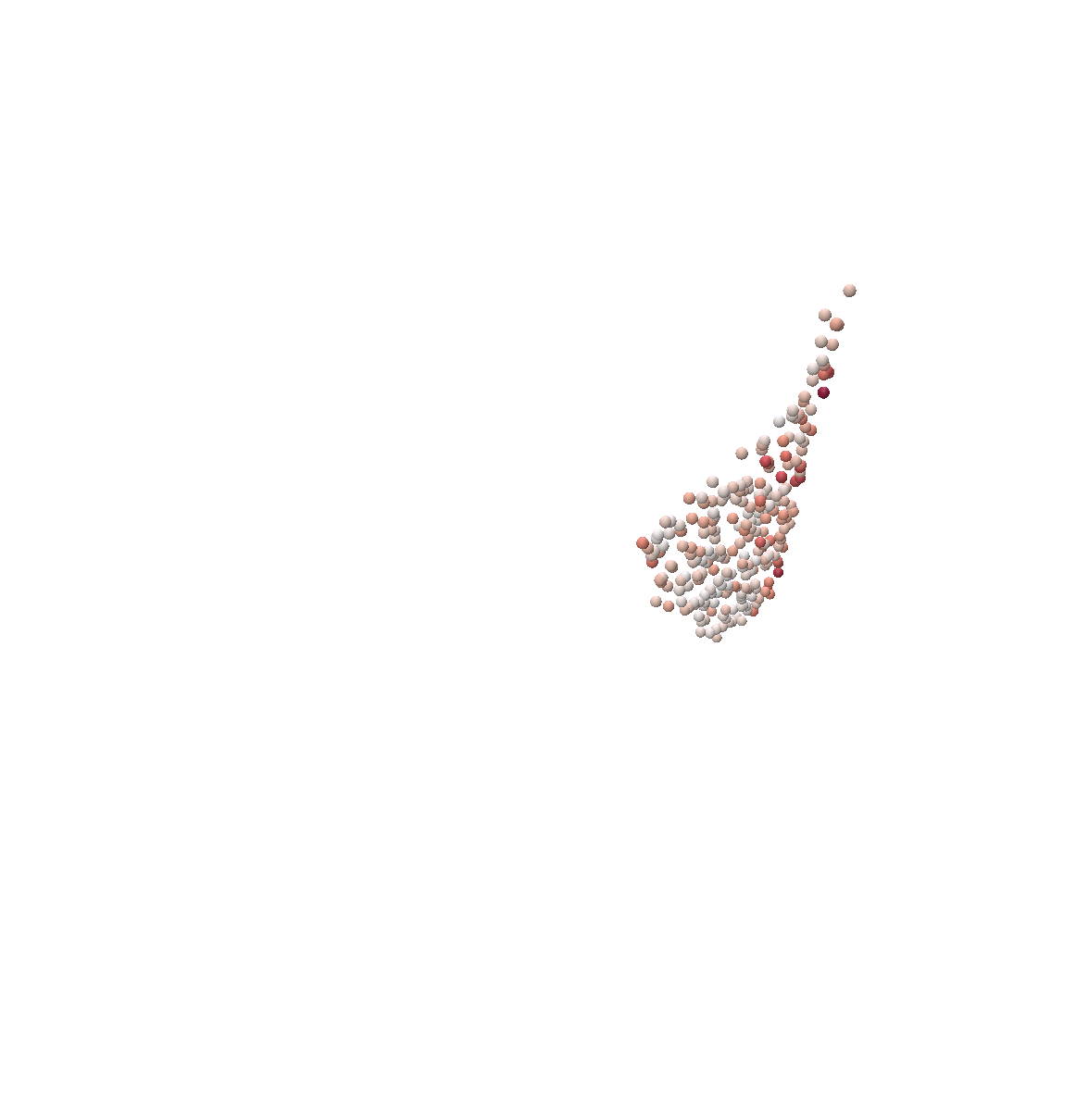} &
        \includegraphics[width=0.18\linewidth]{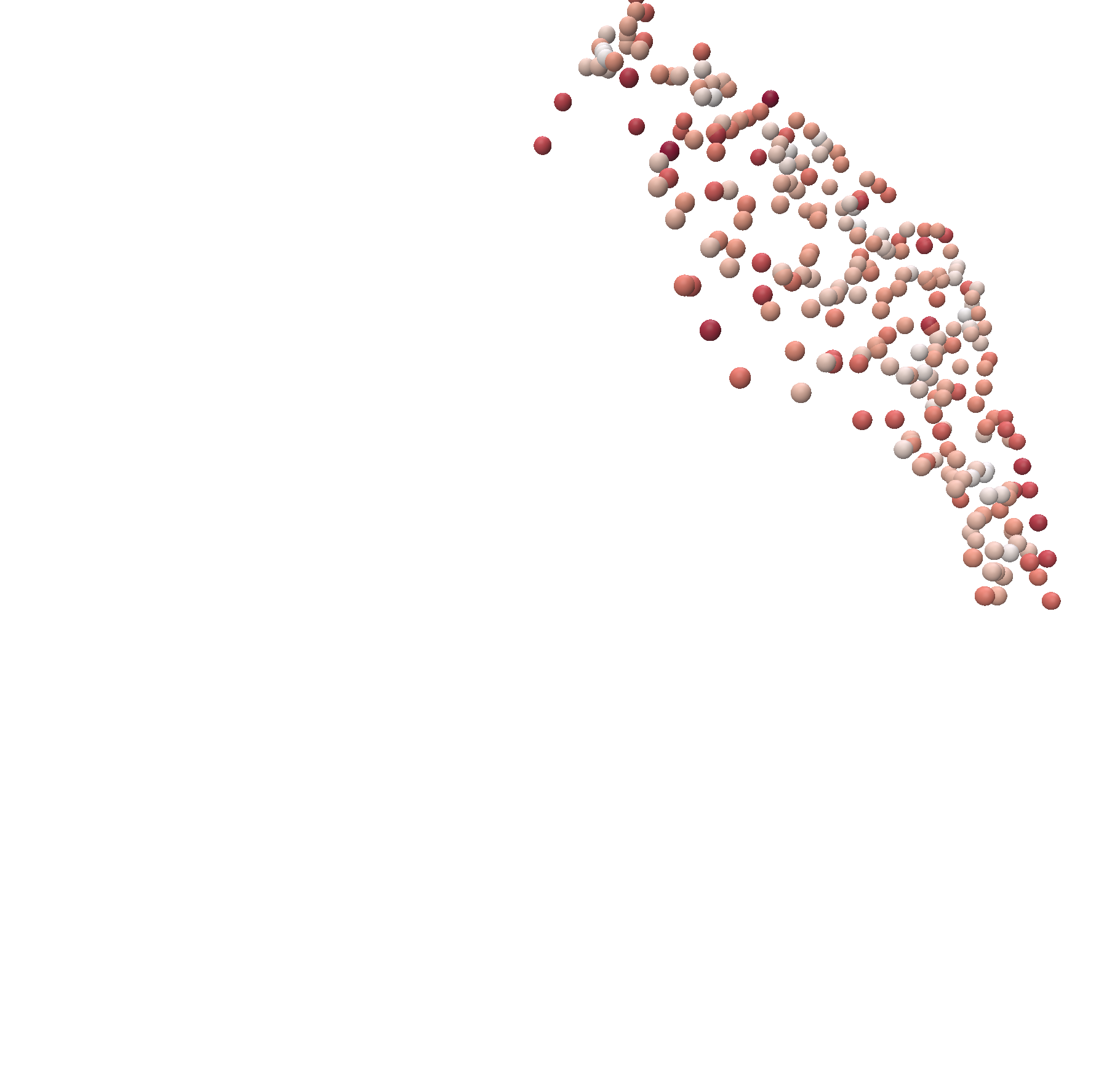} \\
        
        CFPS (Our) & 
        \includegraphics[width=0.13\linewidth]{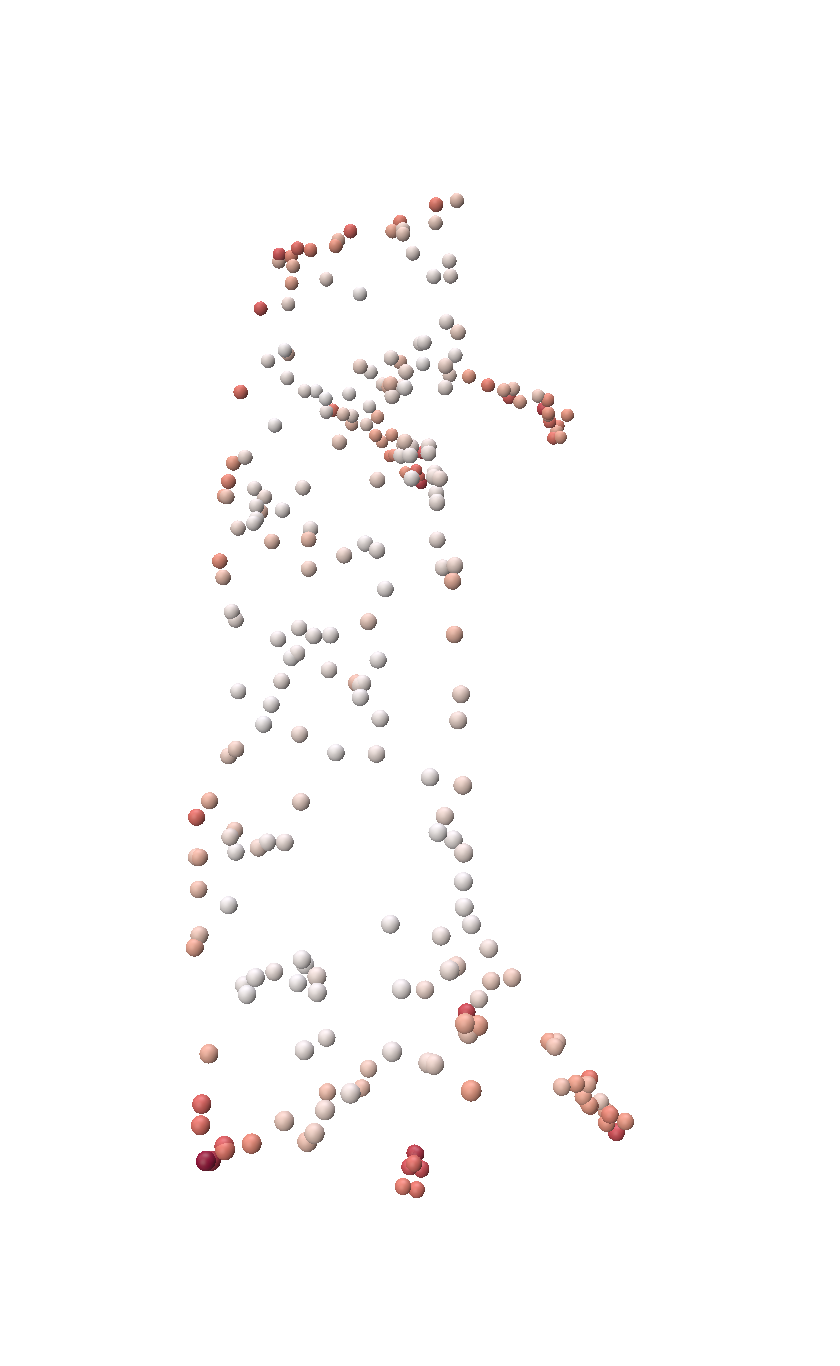} & 
        \includegraphics[width=0.22\linewidth]{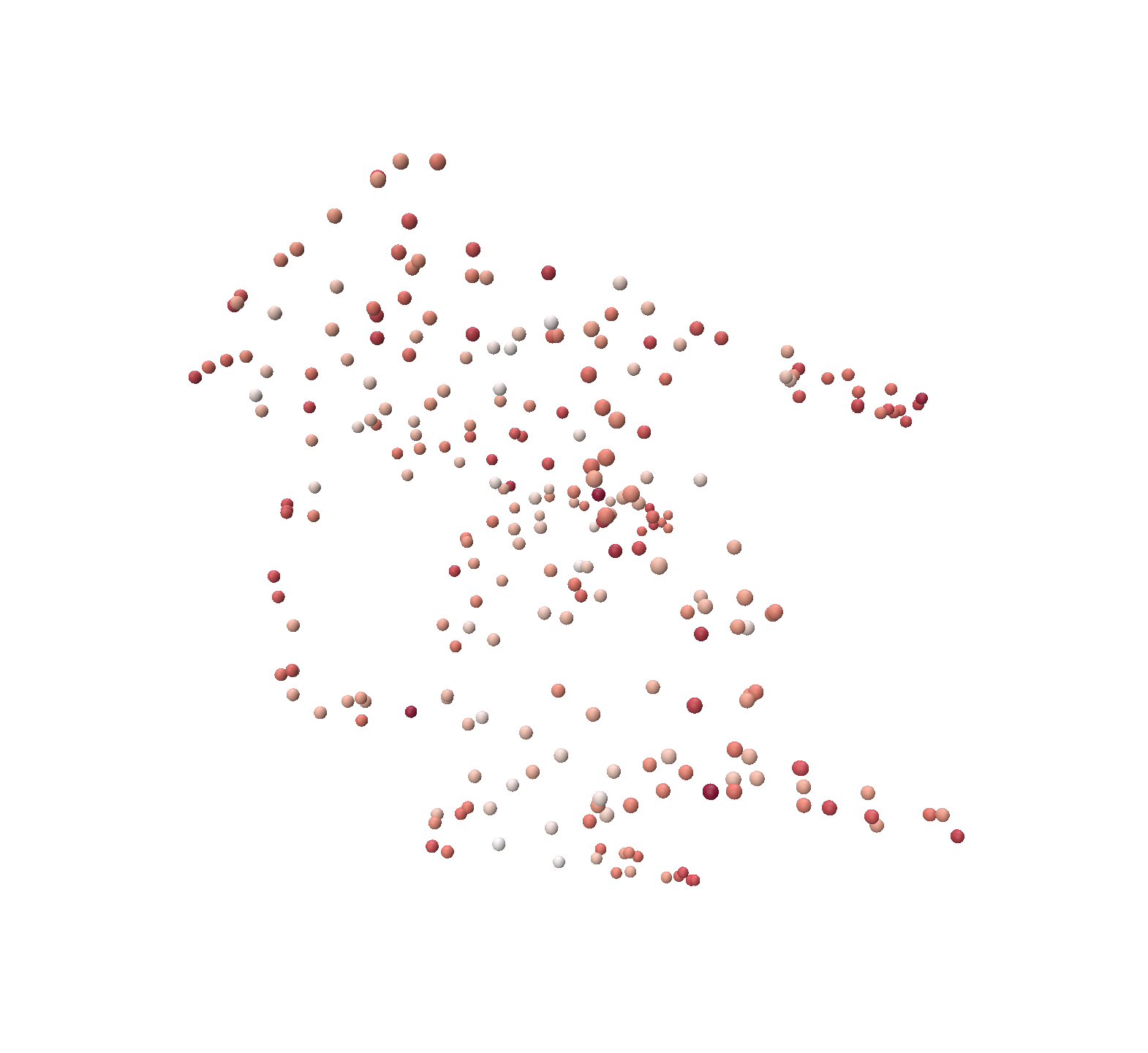} &
        \includegraphics[width=0.22\linewidth]{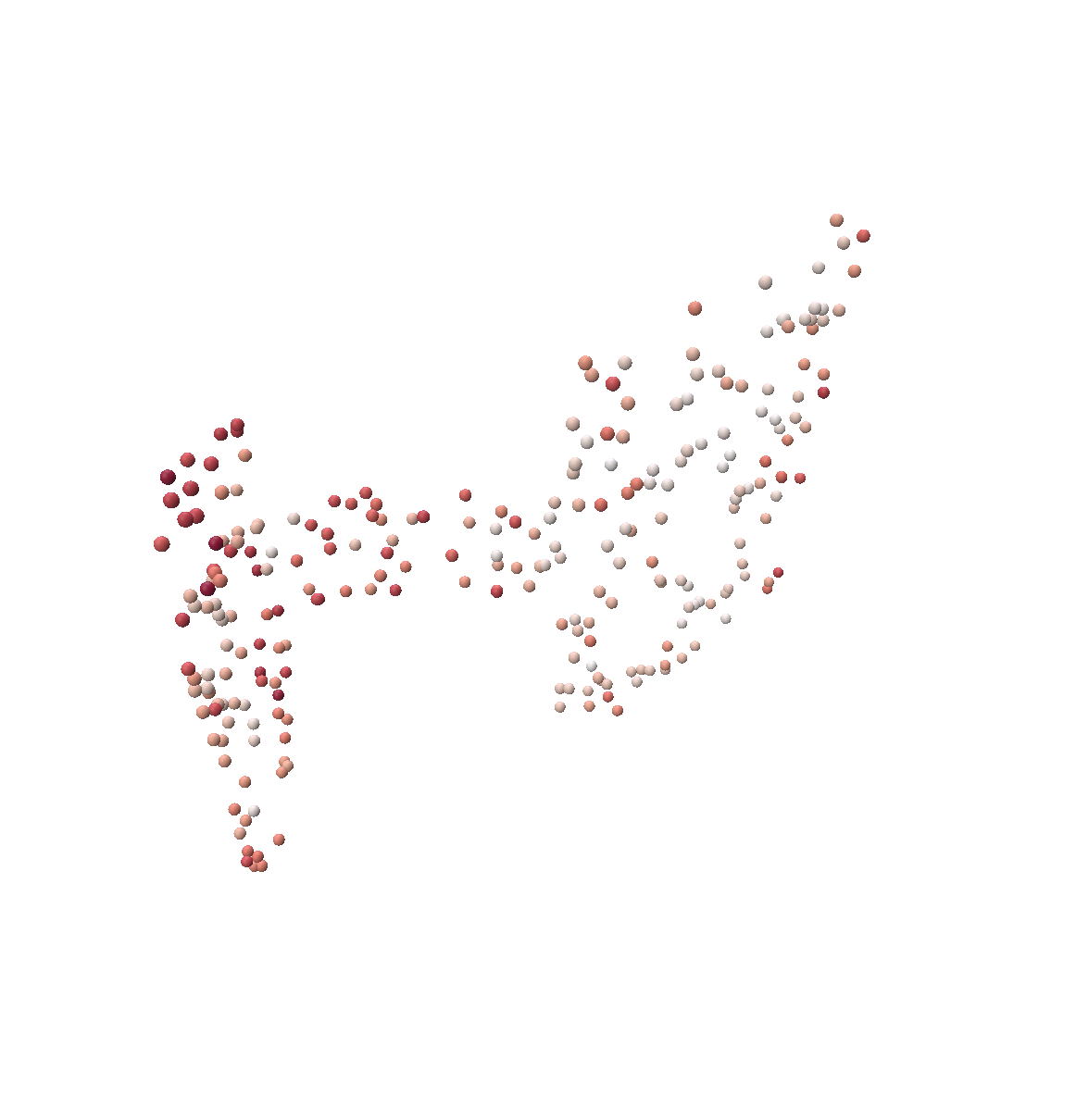} &
        \includegraphics[width=0.18\linewidth]{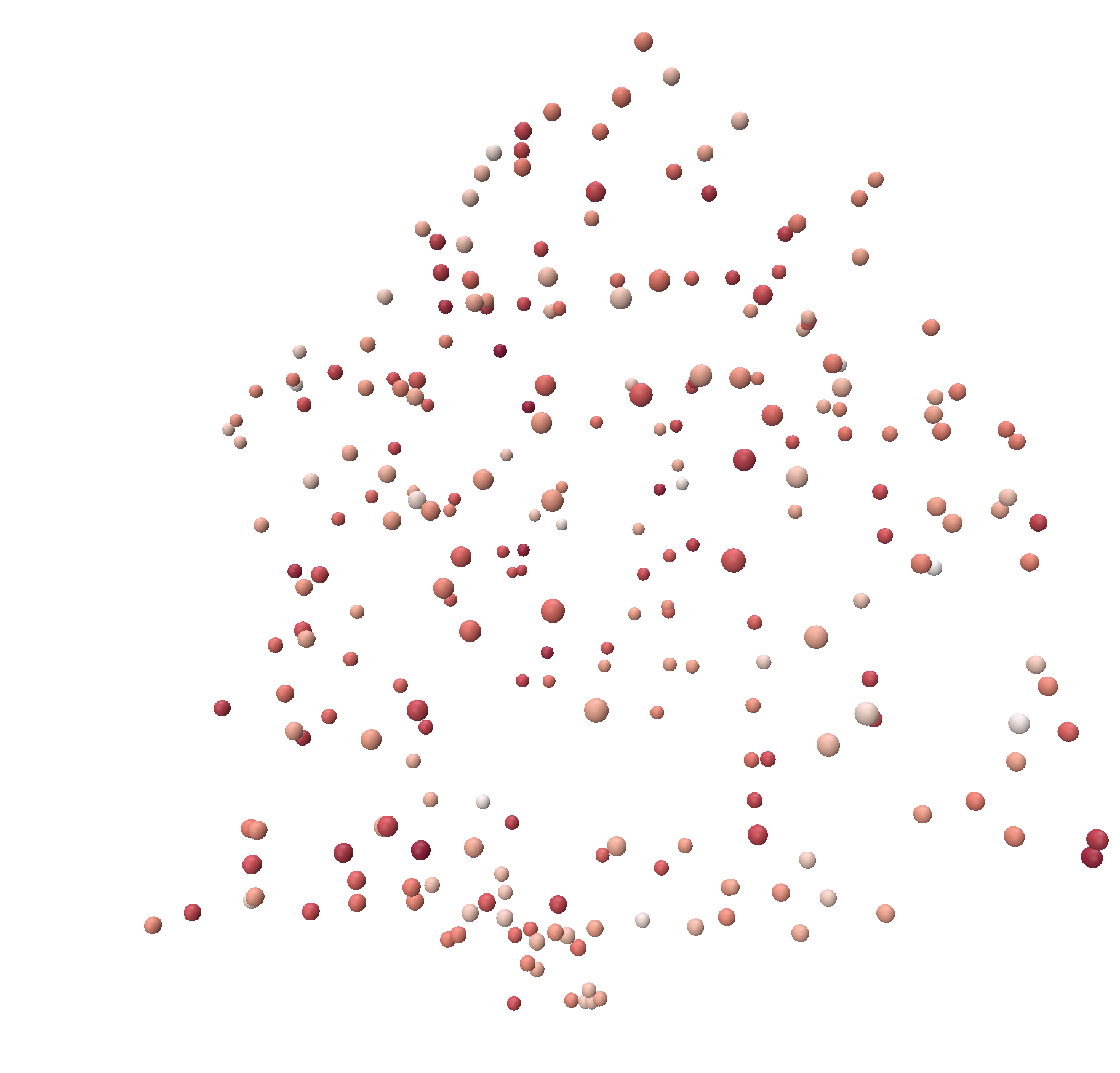} \\
        
    \end{tabular}
    \caption{Comparison of downsampling methods across different objects. Each column represents the point cloud of a specific object (Desk, Chair, Monitor, Dressing Table), either the full point cloud or downsampled. The rows represent the downsampling method: Original Point Cloud, FPS, APES Local, APES Global, and CFPS (Our). The original pointcloud is of size 2048 and is downsampled to 256 points. The scalar map show curvature values increasing from white to red.}
    \label{comparison_table}
\end{figure*}
\subsubsection{Architecture}
For the point cloud completion task, we use the VRCNet architecture \cite{pan2021variational} as the baseline, which has demonstrated strong performance on the MVP dataset. The VRCNet model consists of two main encoder-decoder sub-networks: the PMNet (Probabilistic Modeling Network) and the RENet (Relational Enhancement Network)
\begin{itemize}
    \item \textbf{PMNet:} This module is responsible for generating a coarse, downsampled shape skeleton from the input partial point cloud. This skeleton provides crucial global shape cues for the \textbf{RENet} module.
    \item \textbf{RENet:} The RENet module enhances the structural relations of the point cloud by learning multi-scale point features and reconstructing a high-quality, dense point cloud from the shape skeleton.
\end{itemize}
Refer Supplymentary for simplified architecture diagram.

\subsubsection{Evaluation Metric}
For evaluating the quality of the reconstructed point cloud, we use the \textbf{bi-directional Chamfer Distance} (CD) \cite{fan2017point}, which measures the overall pairwise distance between the nearest points in the predicted and ground truth shapes. However, it has been shown that the CD metric can be misleading in the presence of outliers \cite{tatarchenko2019single}. To address this, we also use the \textbf{F1-Score} \cite{knapitsch2017tanks} to evaluate the distance between the reconstructed and target surfaces. 

\subsubsection{Implementation Details and Results}
We train our model using the default parameters from the official Pytorch Implementation of VRCNet provided in the MVP Benchmark repository. We directly apply our CFPS method to the VRCNet model without modifying its architecture. In our experiments, we also perform ablation studies on the sampling ratio hyperparameter to understand its impact on different object classes presented in the supplementary section.

\subsubsection{Quantitative Results} Table \ref{table:2} compares the performance of different methods on the MVP validation set. The CFPS-based method (VRCNet + CFPS) achieves significant improvements in both Chamfer Distance (CD) and F1-Score compared to other SOTA methods, including PCN \cite{yuan2018pcn}, TopNet \cite{tchapmi2019topnet}, and CRN \cite{wang2020cascaded}. Specifically, CFPS improves the F1-Score from 0.50 with VRCNet + FPS to 0.52, and reduces the CD loss from 5.96 (VRCNet + FPS) to 5.60 (VRCNet + CFPS). A \textit{smaller chamfer distance} ensures that the model fills in the missing regions without introducing broader context or mismatched patterns, making it more local and precise.


\begin{table}[h]
\centering
\begin{tabular}{lcc}
\toprule
\textbf{Method} & \textbf{CD} & \textbf{F1-Score@1\%} \\
\midrule
PCN\cite{yuan2018pcn}        & 9.77   & 0.32   \\
TopNet\cite{tchapmi2019topnet}   & 10.11  & 0.31  \\
CRN\cite{wang2020cascaded}    & 7.25   & 0.43  \\
ECG\cite{pan2020ecg}         & 6.64   & 0.48  \\
VRCNet+FPS                   & 5.96   & 0.50  \\
VRCNet+CFPS                  & \textbf{5.60}    & \textbf{0.52}   \\
\bottomrule
\end{tabular}
\caption{Comparisons of performance with existing methods on the MVP validation set. Both the input and output contain 2048 points. CD loss multiplied by $10^4$.}
\label{table:2}
\end{table}



\subsection{Point Cloud Classification}
Point cloud classification is a fundamental task in 3D computer vision, where the goal is to categorize point cloud data into predefined object classes. For our experiments, we employ the ModelNet10 and ModelNet40 datasets \cite{modelnet} and the PointNet++ architecture to evaluate the performance of our Curvature-aware Furthest Point Sampling (CFPS) method in improving classification results.

\subsubsection{Datasets}
We evaluate our method on two widely used benchmark datasets for 3D object classification: ModelNet10 and ModelNet40. ModelNet10 consists of 10 categories of 3D CAD models, such as chairs, tables, and cars, with a balanced distribution of object types. This dataset is ideal for testing models on relatively simple and structured 3D shapes \cite{wu20153d}. On the other hand, ModelNet40 includes 40 categories and over 12,000 models, covering a broader range of object types including household items, furniture, and vehicles, such as bathtubs, beds, and desks. ModelNet40 provides a more challenging test bed for classification algorithms due to the larger variety and complexity of object geometries.

\subsubsection{Architecture and Metrics}
For classification, we use the PointNet++ architecture \cite{qi2017pointnet++}. PointNet++ builds upon the original PointNet framework by introducing hierarchical feature learning, enabling it to capture local structures at multiple scales. The network uses a set abstraction module that allows for downsampling while maintaining important local geometric features. This makes it particularly suitable for tasks like point cloud classification, where capturing both global and local geometric context is critical. For training details refer supplementary. 

We used the negative log likelihood loss for training and overall accuracy for evaluating our model.



\subsubsection{Quantitative Results} Table \ref{table:1} shows the overall classification accuracy of different sampling methods (FPS, APES, and CFPS) on the ModelNet10 and ModelNet40 datasets. Our CFPS method significantly outperforms both Furthest Point Sampling (FPS) and APES across both datasets, achieving the highest accuracy scores. These results demonstrate the potential of CFPS to improve the overall classification accuracy by selecting more representative points for processing.

\begin{table}[t] 
\centering 
\begin{tabular}{lcc} 
\toprule 
\textbf{Sampling Method} & \textbf{ModelNet10} & \textbf{ModelNet40} \\ 
\midrule 
FPS & 0.990 & 0.97 \\ 
APES (local) & 0.969 & 0.94 \\ 
APES (global) & 0.943 & 0.922 \\ 
CFPS & \textbf{0.996} & \textbf{0.983} \\ 
\bottomrule 
\end{tabular} 
\caption{Overall accuracy of different sampling methods on the ModelNet10 and ModelNet40 datasets. The Curvature-aware Sampling (CFPS) method outperforms both FPS and APES on both datasets, demonstrating its effectiveness.} 
\label{table:1} 
\end{table}

\subsubsection{Qualitative Results}

Figure \ref{comparison_table} shows the point cloud classifications using different sampling methods. The CFPS method leads to more accurate and refined object classifications, especially in complex geometries where APES struggle to capture important details and FPS is not detailed enough at the curvature points. This enables PointNet++ to better learn the distinguishing features of the objects.


\subsection{Ablation Studies}

\textbf{Impact of Exchange Ratio on Different Object Classes for shape generation}: We compare the performance of models with differing exchange ratios, which dictate the proportion of points selected for exchange between the core (FPS-selected) and non-core (remaining) sets.

\begin{itemize}
    \item \textbf{Uniform Curvature Objects} (e.g., Sofas): For objects with generally uniform curvature, the performance remains stable across different exchange ratios. In these cases, the FPS alone sufficiently captures the object’s shape, and the additional curvature-aware ability of CFPS does not provide any significant improvements 
    \item \textbf{High Curvature Variability Objects} (e.g., Beds, Bookshelves, Tables): For objects with a high variability in curvature, the advantages of CFPS become more evident. Increasing the number of exchange points allows the model to capture fine details in regions with sharp edges and complex structures, resulting in improved \textbf{Chamfer Distance (CD)} and \textbf{F1-Score}.
\end{itemize}

A detailed figure for ablation is presented in supplementary.


\textbf{Impact of the Learned Ratio Estimator:} We conduct ablation studies to evaluate the effectiveness of the learned exchange ratio estimator compared to using fixed exchange ratios (e.g., 0.5) for classification and its impact on downstream tasks. \textit{The goal is to assess how an adaptive sampling ratio, learned through our policy gradient approach, influences model performance relative to fixed-ratio baselines.}


\paragraph{Quantitative Comparison:} Table~\ref{tab:ablation_study} summarizes the performance in terms of \textbf{Overall Accuracy} for the learned fixed ratios versus the learned adaptive ratio estimator. 

\begin{table}[h]
    \centering
    \begin{tabular}{lcc}
        \toprule
        \textbf{Ratio} & \textbf{ NLL Loss} & \textbf{Overall Accuracy} \\
        \midrule
        Fixed Ratio (0.1) & 0.0772 & 0.9879 \\
        Fixed Ratio (0.25) & 0.0729 & 0.9945 \\
        Fixed Ratio (0.5) & 0.0674 & 0.9926 \\
        Fixed Ratio (0.75) & 0.0719 & 0.9829 \\
        Fixed Ratio (0.9) & 0.0750 & 0.9967 \\
        \midrule
        Adaptive &  \textbf{0.0660} &  \textbf{0.996} \\
        \bottomrule
    \end{tabular}
    \caption{Performance comparison between fixed-ratio and Adaptive ratio estimators for shape completion. Our Adaptive ratio estimator achieves the highest accuracy and the lowest negative log-likelihood compared to all other fixed-ratio methods.}
    \label{tab:ablation_study}
\end{table}

\textbf{FLOPS and Parameter Overhead:} Table \ref{table:time} shares overhead in FLOPS and paramteres. The pretrained normal estimation ($25.83$M params) adds the major overhead followed by Exchange ratio estimator ($0.04$M params).

\begin{table}[h]
\centering
\resizebox{1\linewidth}{!}{
\begin{tabular}{c|ccccccccccc}
\toprule
Method & S-NET & PST-NET & SampleNet & MOPS-Net & LighTN & APES (local) & APES (global) & CFPS\\  \midrule
Params & 0.33M & 0.42M & 0.46M & 0.44M & 0.37M & 0.35M & 0.35M & 25.87M\\
FLOPs & 152M & 122M & 167M & 149M & 115M & 142M & 114M & 203M\\
\bottomrule
\end{tabular}}
\caption{Computation complexity of different sampling methods. Here \enquote{M} stands for million. \vspace{-0.1cm}} 
\label{table:time}
\end{table}



\section{Conclusion}
We introduce the Curvature informed Furthest Point Sampling algorithm, a variant of the furthest point sampling (CFPS) algorithm derived by integrating curvature into the sampling process and employing a policy network to dynamically adjust the point exchange ratio, we ensure that the final point set captures both spatial coverage and fine geometric detail. \par Our approach balances efficiency with accuracy, improving upon the redundancy often introduced by planar regions in traditional FPS. We showed empirically that our proposed algorithm can achieve better performance on the task of shape completion than the conventional furthest point sampling algorithm.We believe that this framework offers significant potential for advancing point cloud processing in areas such as 3D object recognition, scene reconstruction, and autonomous navigation.






\newpage
\appendix
\section*{Appendix}
\section{Training Strategies}

\subsection{Ablation on Training Strategies}
\begin{table}[h]
\centering
\begin{tabular}{lll}
\toprule
\textbf{Training Strategy} & \textbf{Epochs} & \textbf{F1-Score} \\ 
\midrule
FPS only & 40 & 0.50 \\
CFPS only & 40 & 0.52 \\
FPS + CFPS & 15+25 & 0.507 \\
\bottomrule
\end{tabular}
\caption{Ablation on training strategies}
\label{table:3}
\end{table}

\subsubsection{Impact of Training Strategies:} The ablation study in Table \ref{table:3} compares the performance of different training strategies:
\begin{itemize}
    \item \textbf{FPS-only:} This baseline method uses only Furthest Point Sampling and achieves a F1-Score of 0.50 after 40 epochs of training.
    \item \textbf{CFPS-only:} Using Curvature-aware FPS , the model achieves a F1-Score of 0.52, slightly lower than FPS-only. This indicates that while CFPS improves shape completion accuracy, the fixed sampling ratio is not as effective as an adaptive strategy.
    \item \textbf{FPS + CFPS:} A hybrid approach that first uses FPS to downsample and then applies CFPS for curvature-aware sampling, followed by adaptive sampling through policy gradient. This method achieves the best performance, with a F1-Score of 0.507 after training for 15+25 epochs.
\end{itemize}

This further demonstrates that while CFPS alone helps in improving the sampling quality, the dynamic adjustment of the sampling ratio via policy gradients leads to the most significant gains in performance.

\section{Ablation studies on exchange ratio in shape generation}
\begin{figure*}[h]
\includegraphics[width=\linewidth]{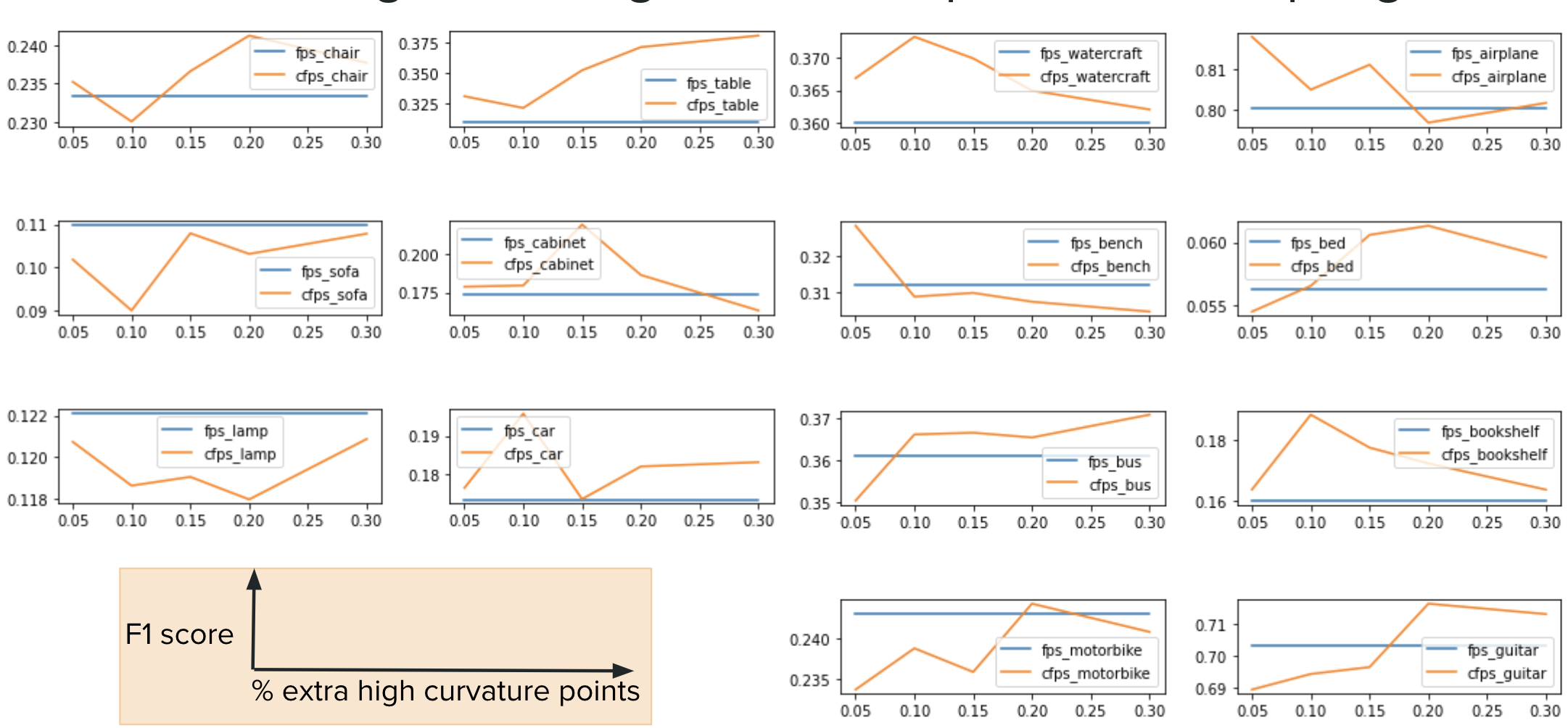}
\caption{Result of changing \protect{$nep$} number of exchange points on different classes in MVP dataset}
\label{fig3}
\end{figure*}

FPS performs as well as CFPS for objects with uniform curvature such as for the example of a sofa. Objects having good curvature variability can see tremendous benefits from using the curvature features such as bed, book-shelf, and table. Refer \cref{fig3} for more details.

\begin{figure*}[h!]
\includegraphics[width=\linewidth]{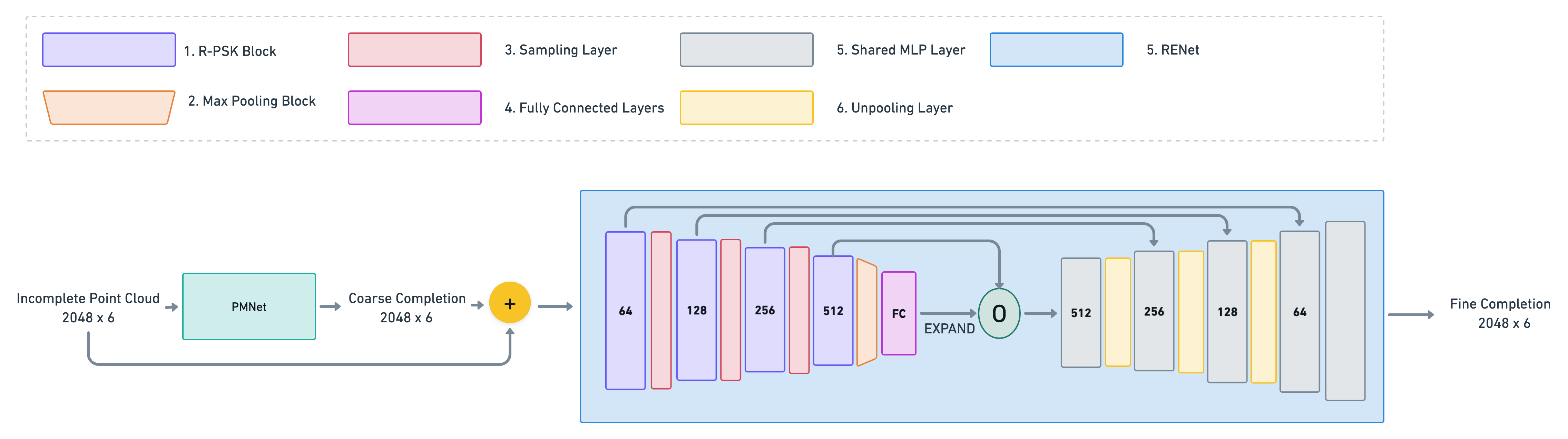}
\caption{VRCNet Architecture for Shape Completion}
\label{fig1}
\end{figure*}

\section{Theoretical Regret Bounds}
We derive high probability regret bounds for the REINFORCE algorithm applied to curvature-aware furthest point sampling (CFPS). This analysis extends classic reinforcement learning theory by applying it to a custom point cloud sampling problem. 
\subsection{Regret Formulation:} We define the cumulative regret after $T$ time steps, $R_T$, which is the difference between the expected rewards under the optimal policy $\pi^*$ and the learned policy $\pi_\phi$ parameterized by $\phi$:
\begin{equation}
R_T = \sum_{t=1}^{T} \left( \mathbb{E}_{\pi^*}[\mathcal{R}(s_t, a_t)] - \mathbb{E}_{\pi_\phi}[\mathcal{R}(s_t, a_t)] \right)
\end{equation}
where $\mathcal{R}(s_t, a_t) = - \mathcal{L}_\phi(X, G_t)$ 
\subsection{REINFORCE Gradient Update:}

Let $J(\phi) = \mathbb{E}_{\pi_\phi} \left[ \mathcal{R}(s_t, a_t) \right]$. Using the REINFORCE algorithm, we perform gradient updates of the policy parameters $\phi$ based on the expected reward \cite{sutton2018reinforcement}:
\begin{equation}
\phi_{t+1} = \phi_t + \alpha_t \nabla_\phi J(\phi)
\end{equation} Using the log-likelihood trick \cite{williams1992simple}, we get the gradients to reduce to:
\begin{equation}
\nabla_\phi J(\phi) = \mathbb{E}_{\pi_\phi} \left[ \nabla_\phi \log \pi_\phi(a_t | s_t) \mathcal{R}(s_t, a_t) \right]
\end{equation}
 .
\subsection{Assumptions and Conditions:} We make the following assumptions while deriving the regret bound:
\begin{itemize}
    \item \textbf{Bounded rewards:} There exists a constant $M > 0$ such that $|\mathcal{R}(s_t, a_t)| \leq M\ \forall\ t$ \cite{shalev2014understanding}.
    \item \textbf{Lipschitz continuity:} There exists a constant $L > 0$ such that for any two policies $\pi_1$ and $\pi_2$ \cite{virmaux2018lipschitz}:
    \begin{equation}
    |\mathcal{R}(s_t, a_t; \pi_1) - \mathcal{R}(s_t, a_t; \pi_2)| \leq L \|\phi_1 - \phi_2\|
    \end{equation} \par 
    We can show that our system is locally Lipschitz continuous if the conditions hold:
    \begin{enumerate}
        \item \textbf{Weight norms:} If the weights of the neural network are bounded, local Lipschitz continuity holds for the current parameters.
        \item \textbf{Gradient control:} By establlishing the gradient norms during training, we can ensure that updates to the policy parameters do not lead to drastic changes in the output probabilities, which satisfies a local Lipschitz condition \cite{gouk2021regularisation}.
        \item \textbf{Finally}, We use a simple MLP architecture which is more conducive to achieving tighter regret bounds in our problem.
    \end{enumerate}
    
\end{itemize}

\begin{table}[t]
\centering
\scalebox{0.82}{
\begin{tabular}{lc}
\toprule
Method & Overall Accuracy \\ \midrule
PointNet \cite{li2018pointcnn} & 89.2\% \\
PointNet++ \cite{qi2017pointnet++} & 91.9\% \\
SpiderCNN \cite{Xu_2018_ECCV} & 92.4\% \\
DGCNN \cite{wang2019dynamicgraphcnnlearning} & 92.9\% \\
PointCNN \cite{li2018pointcnn} & 92.2\% \\
PointConv \cite{Wu_2019_CVPR} & 92.5\% \\
PVCNN \cite{NEURIPS2019_57370345} & 92.4\% \\
KPConv \cite{thomas2019kpconv} & 92.9\% \\
PointASNL \cite{yang2020teaser} & 93.2\% \\
PT$^1$ \cite{9552005} & 92.8\% \\
PT$^2$ \cite{Zhao_2021_ICCV} & 93.7\% \\
PCT \cite{guo2021pct} & 93.2\% \\
PRA-Net \cite{cheng2021net} & 93.7\% \\
PAConv \cite{xu2021paconv}  & 93.6\% \\
CurveNet \cite{muzahid2020curvenet} & 93.8\% \\
DeltaConv \cite{wiersma2022deltaconv} & 93.8\% \\ \midrule
APES (local-based) & 93.5\% \\
APES (global-based) & 93.8\% \\
FPS & 94.8\% \\
CFPS & \textbf{96.2\%} \\ \bottomrule
\end{tabular}}
\caption{Classification results on ModelNet40. In comparison with other SOTA methods that also only use raw point clouds as input. Note that our reported results did not consider the voting strategy.}
\label{table:cls}
\end{table}

\begin{table}[t]
\centering
\resizebox{1\linewidth}{!}{
\begin{tabular}{c|ccccccccccc}
\toprule
Method & S-NET & PST-NET & SampleNet & MOPS-Net & LighTN & APES (local) & APES (global) & CFPS\\  \midrule
Params & 0.33M & 0.42M & 0.46M & 0.44M & 0.37M & 0.35M & 0.35M & 25.87M\\
FLOPs & 152M & 122M & 167M & 149M & 115M & 142M & 114M & 203M\\
\bottomrule
\end{tabular}}
\caption{Computation complexity of different sampling methods. Here \enquote{M} stands for million. \vspace{-0.1cm}} 
\label{table:time}
\end{table}

\subsection{Hoeffding's Bounds:} 
Hoeffding’s inequality \cite{hoeffding1994probability}  for bounded random variables $X_1, \dots, X_T$ implies:
\begin{equation}
\mathbb{P} \left( \sum_{t=1}^{T} (X_t - \mathbb{E}[X_t]) \geq \epsilon \right) \leq \exp\left( \frac{-2 \epsilon^2}{T (b-a)^2} \right)
\end{equation}

where $X_t = \mathbb{E}_{\pi^*}[\mathcal{R}(s_t, a_t)] - \mathbb{E}_{\pi_\phi}[\mathcal{R}(s_t, a_t)]$


 Since rewards are bounded by $M$,
 applying Hoeffding’s inequality with $a = 2M$ and $b = 0$ and  $X_t$ is i.i.d then $\mathbb{E}[X_t] = 0$   gives:
\begin{equation}
\mathbb{P} \left( \sum_{t=1}^{T} X_t  \geq \epsilon \right) \leq \exp\left( \frac{-2 \epsilon^2}{T (2M)^2} \right)
\end{equation}
Setting the probability of exceeding $\epsilon$ to $\delta$:
\begin{equation}
\mathbb{P} \left( R_T \geq \epsilon \right) \leq \delta
\end{equation} we solve for $\epsilon$:
\begin{equation}
\epsilon \leq M \sqrt{2T \log(1/\delta)}
\end{equation} Thus, with probability at least $1 - \delta$, the regret $R_T$ is bounded by \cite{shalev2014understanding}:
\begin{equation}
R_T \leq M \sqrt{2 T \log(1/\delta)}
\end{equation}
Therefore, we derived a high probability regret bound for the REINFORCE algorithm applied to CFPS, showing that the regret grows sublinearly as $\mathcal{O}(\sqrt{T \log(1/\delta)})$. 

\section{Training Details}
To train the model, we use the Adam Optimizer with a learning rate of $1 \times 10^{-4}$ and set a learning rate of $2 \times 10^{-2}$ for the REINFORCE agent. We train the algorithm with a batch size of 4 for 100 epochs on a single A100 GPU.

\subsubsection{Training Details - Classification} 
We train the PointNet++ model with our CFPS-based sampling technique using the default training settings from the official PointNet++ repository. The model is trained with the negative log-likelihood loss function \cite{lecun2015deep}, which is commonly used for classification tasks in deep learning. 

For the training procedure, we use the Adam optimizer with an initial learning rate of $0.001$. The model is trained for 10 epochs, with a batch size of 8 using NVIDIA A100 GPU.

\subsection{Point Cloud Segmentation}
Point cloud segmentation is a fundamental task in 3D computer vision, where the objective is to assign semantic labels to points in a 3D space, enabling fine-grained understanding of object parts. This section evaluates the performance of different downsampling strategies—FPS, APES, and CFPS—integrated into the Inductive Bias-Aided Transformer (IBT) framework on the ShapeNetPart dataset.

\subsubsection{Dataset}
We conduct experiments on the \textbf{ShapeNetPart} dataset \cite{yi2016scalable}, which is widely used for 3D part segmentation tasks. The dataset comprises 16,880 3D models across 16 object categories, annotated with 50 semantic part labels. The dataset is split into 14,006 models for training and 2,874 for testing. Each object contains between 2 and 6 annotated parts, representing diverse geometric and semantic challenges. The input to all models is uniformly sampled point clouds with 1,024 points per object.

\subsubsection{Experimental Setup and Metrics}
To evaluate the impact of downsampling methods, we incorporate FPS, APES (local and global), and CFPS as downsampling layers into the IBT framework. The IBT model processes the input point clouds through its key components:
\begin{itemize}
    \item \textbf{Relative Position Encoding}: Encodes spatial relationships and geometric structures within local neighborhoods.
    \item \textbf{Attentive Feature Pooling}: Aggregates local features using a combination of max pooling and attention mechanisms.
    \item \textbf{Locality-Aware Transformer}: Enhances global self-attention by incorporating local spatial features through channel-wise modulation.
\end{itemize}

We use the following evaluation metrics:
\begin{itemize}
    \item \textbf{Mean Intersection over Union (mIoU) per class}: Average IoU across all semantic categories.
    \item \textbf{Mean Intersection over Union (mIoU) per shape}: Average IoU across all objects, irrespective of category.
\end{itemize}


\subsubsection{Quantitative Results}
Table \ref{table:segmentation_results} shows the segmentation performance of FPS, APES (local and global variants), and CFPS. CFPS achieves the highest mIoU in both class-level and shape-level metrics, highlighting its ability to preserve critical geometric features essential for accurate segmentation.



\begin{table}[t]
\centering
\scalebox{0.82}{
\begin{tabular}{lcc}
\toprule
Method & Cat. mIoU & Ins. mIoU \\ \midrule
PointNet \cite{qi2017pointnet} & 80.4\% & 83.7\% \\
PointNet++ \cite{qi2017pointnet++} & 81.9\% & 85.1\% \\
SpiderCNN \cite{Xu_2018_ECCV} & 82.4\% & 85.3\% \\
DGCNN \cite{wang2019dynamicgraphcnnlearning} & 82.3\% & 85.2\% \\
SPLATNet \cite{su2018splatnet} & 83.7\% & 85.4\% \\
PointCNN \cite{li2018pointcnn} & 84.6\% & 86.1\% \\
PointConv \cite{Wu_2019_CVPR} & 82.8\% & 85.7\% \\
KPConv \cite{thomas2019kpconv} & 85.0\% & 86.2\% \\
PT$^1$ \cite{9552005} & - & 85.9\% \\
PT$^2$ \cite{Zhao_2021_ICCV} & 83.7\% & 86.6\% \\
PCT \cite{guo2021pct} & - & 86.4\% \\
PRA-Net \cite{cheng2021net} & 83.7\% & 86.3\% \\
PAConv \cite{xu2021paconv} & 84.6\% & 86.1\% \\
CurveNet \cite{muzahid2020curvenet} & - & \textbf{86.6\%} \\
StratifiedTransformer \cite{lai2022stratified} & 85.1\% & 86.6\% \\
\midrule
APES (local-based) & 83.1\% & 85.6\% \\
APES (global-based) & 83.7\% & 85.8\% \\
FPS & 83.0\%& 83.0\% \\
CFPS & \textbf{84.5\%} & \textbf{86.7\%} \\ \bottomrule
\end{tabular}}
\caption{Segmentation results on ShapeNet Part.}
\label{table:segmentation_results}
\end{table}




\end{document}